\documentclass{siamonline220329}

\usepackage{amsmath}
\usepackage{amsfonts}
\usepackage{amssymb}
\usepackage{mathtools}
\usepackage{bbm}
\usepackage{multirow}
\usepackage{graphicx}
\usepackage[small]{caption}
\usepackage{subcaption}
\usepackage{epstopdf}
\usepackage[shortlabels]{enumitem}
\setlist[enumerate]{leftmargin=.5in}
\setlist[itemize]{leftmargin=.3in}

\usepackage[margin=1in]{geometry}

\usepackage{algorithm}
\usepackage{algorithmic}
\usepackage{bm}
\usepackage{underoverlap}

\usepackage[titletoc]{appendix}

\usepackage[nameinlink]{cleveref}
\crefname{equation}{}{} 

\newtheorem{thm}{Theorem}[section]

\newtheorem{lem}[thm]{Lemma}
\newtheorem{prop}[thm]{Proposition}

\numberwithin{equation}{section}

\DeclareMathOperator*{\argmax}{arg\, max}

\DeclareMathOperator{\diag}{diag}

\newcommand{\se}{\text{e}}

\newcommand{\bR}{\mathbb{R}}

\newcommand{\bE}{\mathbb{E}}
\newcommand{\bN}{\mathbb{N}}

\newcommand{\vect}[1]{\bm{#1}}

\newcommand{\vc}{\vect{c}}

\newcommand{\vs}{\vect{s}}

\newcommand{\xbm}{\vect{x}}
\newcommand{\vu}{\vect{u}}
\newcommand{\vx}{\vect{x}}
\newcommand{\vy}{\vect{y}}

\newcommand{\vgamma}{\vect{\gamma}}
\newcommand{\vepsilon}{\vect{\epsilon}}
\newcommand{\vlambda}{\vect{\lambda}}
\newcommand{\vmu}{\vect{\mu}}
\newcommand{\vtheta}{\vect{\theta}}
\newcommand{\vzero}{\vect{0}}

\newcommand{\cN}{\mathcal{N}}

\newcommand{\mtrx}[1]{\mathsf{#1}}

\newcommand{\mA}{\mtrx{A}}
\newcommand{\mB}{\mtrx{B}}
\newcommand{\mC}{\mtrx{C}}

\newcommand{\mF}{\mtrx{F}}
\newcommand{\mH}{\mtrx{H}}
\newcommand{\mI}{\mtrx{\mathrm{I}}}

\newcommand{\mS}{\mtrx{S}}

\newcommand{\mZ}{\mtrx{Z}}

\newcommand{\mGamma}{\mtrx{\Gamma}}

\newcommand{\mSigma}{\mtrx{\Sigma}}

\newcommand{\tg}{\tilde{g}}

\usepackage{todonotes}

\def\Ncal{{\mathcal{N}}}

\def\Rbb{{\mathbb{R}}}

\newcommand{\mubm}{\bm{\mu}}

\def\xbm{{\bm{x}}}
\def\ybm{{\bm{y}}}

\DeclareMathOperator{\EM}{EM}
\DeclareMathOperator{\CB}{CB}
\DeclareMathOperator{\MK}{MK}

\DeclareMathOperator{\SCA}{SCA}
\DeclareMathOperator{\SNR}{SNR}
\DeclareMathOperator{\SQ}{SQ}

\title{Hyperparameter Estimation for Sparse Bayesian Learning Models
\thanks{The work of G. Song was supported in part by the National Science Foundation under grant DMS-1939203. The work of L. Shen was supported in part by the National Science Foundation under the grant DMS-2208385.}
}

\author{ 
    Feng Yu\thanks{Department of Mathematics, The University of Minnesota, Minneapolis, MN, USA (\email{fyu@umn.edu})}
    \and
    Lixin Shen\thanks{Department of Mathematics, Syracuse University, Syracuse, NY, USA (\email{lshen03@syr.edu})}
    \and
    Guohui Song\thanks{Department of Mathematics and Statistics, Old Dominion University, Norfolk, VA, USA (\email{gsong@odu.edu})}
}

\begin{document}

\maketitle

\begin{abstract}
Sparse Bayesian Learning (SBL) models are extensively used in signal processing and machine learning for promoting sparsity through hierarchical priors. The hyperparameters in SBL models are crucial for the model's performance, but they are often difficult to estimate due to the non-convexity and the high-dimensionality of the associated objective function. This paper presents a comprehensive framework for hyperparameter estimation in SBL models, encompassing well-known algorithms such as the expectation-maximization (EM), MacKay, and convex bounding (CB) algorithms. These algorithms are cohesively interpreted within an alternating minimization and linearization (AML) paradigm, distinguished by their unique linearized surrogate functions. Additionally, a novel algorithm within the AML framework is introduced, showing enhanced efficiency, especially under low signal noise ratios. This is further improved by a new alternating minimization and quadratic approximation (AMQ) paradigm, which includes a proximal regularization term. The paper substantiates these advancements with thorough convergence analysis and numerical experiments, demonstrating the algorithm's effectiveness in various noise conditions and signal-to-noise ratios.
\end{abstract}



\section{Introduction}
Bayesian models are pervasive in a wide variety of fields and have numerous applications, including machine learning \cite{gelman1995bayesian, Bishop2006, Barber2012}, signal/image processing \cite{Candy2016, Babacan2010a, Geman1986}, and inverse problems \cite{Calvetti2007a, Calvetti2018a}. Comparing with the classical regularization methods, Bayesian models have the advantage of providing a probabilistic framework for uncertainty quantification. The Bayesian approach provides a natural way to incorporate prior knowledge into the model, which is often useful in practice. In particular, hierarchical Bayesian models provide a flexible framework for incorporating prior knowledge into the model. On the other hand, sparsity is a common property of many real-world problems. For example, in signal processing, the signal of interest is often sparse in some domain, such as the wavelet domain \cite{Donoho2003, Donoho2006}. In machine learning, the sparsity of the model is often desirable for interpretability and computational efficiency \cite{Hastie2015}. The sparse Bayesian learning (SBL) models \cite{wipf2006bayesian, wipf2009unified,Faul2001, Tipping2001a,Wipf2004a} are hierarchical Bayesian models that have been widely used in signal processing and machine learning to promote sparsity through hierarchical priors.

The SBL models operate on hierarchical Bayesian frameworks, encompassing two tiers of parameters: the unidentified signals/weights and the hierarchical prior's hyperparameters.  These hyperparameters are crucial in SBL models, dictating the weights' sparsity  and influencing its performance. In particular, an individual hyperparameter is associated independently with each weight, allowing variance in the magnitudes. This individualized prior formulation is a key aspect of SBL models \cite{Tipping2001a}, as it enables the model to autonomously determine the sparsity pattern of the weights. However, it introduces extra hyperparameters whose dimension is equal to the number of weights, which can be prohibitively large. This creates a significant challenge in the application of SBL models, as the hyperparameters must be estimated from the data. 

A common approach of hyperparameter estimation is to use the empirical Bayes approach (or Type II maximum likelihood) \cite{Bishop2006,murphy2022probabilistic}, which computes the marginal likelihood through integrating out the unknown weights and then maximizes this marginal likelihood with respect to the hyperparameters. This approach is also known as the evidence maximization approach \cite{MacKay1992, MacKay1992a, MacKay1999, Tipping2001a}. However, this marginal likelihood function is often non-convex. That is, we need to solve a high-dimensional non-convex optimization problem to estimate the hyperparameters. It is necessary to develop efficient optimization algorithms to solve this challenging minimization problem.

Various algorithms have been proposed to address this issue, including the expectation-maximization (EM) algorithm \cite{Dempster1977, wipf2007empirical}, the MacKay (MK) algorithm \cite{MacKay1992, MacKay1992a, MacKay1999, Tipping2001a}, and the convex bounding (CB) algorithm \cite{Wipf2007a}. The EM algorithm has guaranteed convergence \cite{Wu1983, Redner1984}, but its convergence rate is often slow \cite{Redner1984}, and it exhibits sensitivity to initial values \cite{Ma2005}. The MK algorithm has been observed to have a faster convergence in many applications \cite{Tipping2001a}, but it currently lacks a theoretical guarantee for convergence. The CB algorithm \cite{Wipf2007a} has a convergence rate comparable to that of the MK algorithm and convergence guarantees. It is not clear how to compare the theoretical convergence rates of these algorithms and how to improve their convergence rates.

In this paper, we present a unified framework that encapsulating these algorithms used for hyperparameter estimation in SBL models. Specifically, we will show that these methods could be cohesively understood within an alternating minimization and linearization (AML) paradigm, distinguished only by their choices of linearized surrogate functions. This integrative framework offers a new perspective through which to interpret, analyze, and compare these algorithms, both theoretically and numerically. In addition, we introduce a novel algorithm through a different choice of the linearized surrogate function, which exhibits superior efficiency compared to its predecessors when the signal noise ratio is low. We further propose a new alternating minimization and quadratic approximation (AMQ) paradigm to improve its performance through adding a proximal regularization term into the proposed linearized surrogate function. We underpin our claims with rigorous convergence analysis and numerical experiments, demonstrating the potency of the proposed algorithm under various scenarios with different noise levels and signal-to-noise ratios. 

While a framework akin to ours has been introduced in \cite{hashemi2021unification} for the three mentioned algorithms, our approach stands distinct in several key aspects. The work in \cite{hashemi2021unification} interprets these algorithms through the lens of majorization-minimization (MM), employing varied techniques to derive the majorants. However, it remains ambiguous which techniques truly enhance efficiency and convergence. In constrast, our perspective is more nuanced. We see surrogate functions as linear approximations of the objective function concerning different change of variables. This viewpoint not only simplifies theoretical comparisons, as showcased in our convergence rate analysis in \Cref{sec:denoising}, but also facilitates the creation of more efficient algorithms. In addition, our method allows for a broader selection of surrogate functions. Unlike \cite{hashemi2021unification} where the surrogate must majorize the objective function, our surrogates aren't bound by this constraint, granting us greater flexibility and potential for algorithmic improvements.

This paper is structured as follows. In \Cref{sec:SBL_intro}, we will present an introduction to the SBL models and discuss the problem of hyperparameter estimation. In \Cref{sec:MM_framework}, we will introduce the AML framework and illustrate how existing algorithms for hyperparameter estimation in SBL models can be redefined within the AML framework. \Cref{sec:denoising} will be dedicated to the proposal of a novel algorithm grounded in the AML framework, with a comparative evaluation of its performance against established algorithms in the denoising scenario. In \Cref{sec:general}, we will further propose a new AMQ approach to improve the proposed AML algorithm, accompanied by a comprehensive analysis of its convergence. \Cref{sec:numerical} will be devoted to presenting the results of numerical experiments designed to showcase the efficacy of the proposed algorithm. Finally, we will summarize our findings and draw conclusions in \Cref{sec:conclusion}.

\section{Sparse Bayesian Learning Models}\label{sec:SBL_intro}
In this section, we will introduce the Sparse Bayesian Learning (SBL) model \cite{Faul2001,Tipping2001a,Wipf2007a}.  Specifically, we consider the following linear inverse problem:
\begin{align}\label{eq:general_problem}
    \vy = \mF\vx + \vepsilon,
\end{align}
where $\mF\in\Rbb^{m\times n}$ is the given dictionary of features, \(\vx \in \bR^{n} \) is the vector of unknown weights, \(\vepsilon \in \bR^{m} \) is the vector of noises, and \(\vy\in \bR^{m} \) is the observation vector. We assume that the noise vector \(\vepsilon\) follows an independent and identically distributed (i.i.d.) normal distribution with zero mean and inverse variance \(\beta\), i.e., \(\vepsilon\sim\Ncal(0,\beta^{-1}\mI)\), where \(\mathcal{N} \) denotes the multivariate normal distribution. That is, the likelihood function is given by
\begin{align}\label{eq:likelihood}
    p(\vy|\vx) = \cN(\vy|\mF\vx, \beta^{-1}\mI).
\end{align}
We also assume the unknown vector \(\vx \) has the following \emph{prior}  distribution:
\begin{align}\label{eq:prior}
    p(\vx|\vgamma) = \cN(\vx|\vzero,\mGamma),
\end{align}
where \(\vgamma\in \bR_{+}^{n}: =  \{\vc\in \bR^{n}: c_{i} \geq 0, 1 \leq  i\leq n\}  \) collects all the \emph{hyperparameters} \(\gamma_{i}\), and \(\mGamma = \diag(\vgamma) \) is the diagonal matrix with diagonal entries given by \(\vgamma\). 

A key aspect of SBL lies in its assumption of distinct hyperparameters \(\gamma_{i}\) for each weight \(x_{i}\), as opposed to employing a common hyperparameter \(\gamma\) for all \(x_{i}\)'s. This individualized approach is crucial for the model's effectiveness. Each hyperparameter \(\gamma_{j}\) specifically governs the sparsity of its corresponding weight \(x_{j}\) in the sense that as \(\gamma_{j}\) approaches zero, it increasingly suggests that the weight \(x_{j}\) is likely to be zero. In the extreme scenario where \(\gamma_{j} = 0\), the weight \(x_{j}\) essentially reduces to a degenerate distribution, represented as a point mass at zero. The capacity of the SBL model to allow for distinct \(\gamma_{i}\)'s, which are learned from data, enables it to autonomously determine the sparsity pattern of the weights \(\vx\). This approach effectively eliminates the need for manual hyperparameter tuning, a significant advantage in practical applications.

We next introduce how to estimate the hyperparameters \(\vgamma \) in the SBL model. The evidence maximization (Type II maximum likelihood) approach \cite{MacKay1992, MacKay1992a, MacKay1999, Tipping2001a} will be employed to estimate the hyperparameters \(\vgamma \) through maximizing the marginal likelihood (evidence) function:
\begin{align}\label{eq:max_evidence}
    \hat{\vgamma} = \argmax_{\vgamma\in \bR_{+}^{n}} p(\vy|\vgamma).
\end{align}
Since both the likelihood \( p(\vy|\vx)\) \cref{eq:likelihood} and the prior \(p(\vx|\vgamma)\) \cref{eq:prior} are Gaussian, it follows from the conjugate property of Gaussian distribution~\cite{murphy2022probabilistic} that the evidence function is given by
\begin{align}\label{eq:evidence}
    p(\vy|\vgamma) = \int  p(\vy|\vx) p(\vx|\vgamma) d\vx  = \cN(\vy|\vzero, \mS(\vgamma)), \quad \mbox{with }  \mS(\vgamma) = \beta^{-1}\mI + \mF\mGamma\mF^{\mathsf{T}}.
\end{align}
Substituting \cref{eq:evidence} into \cref{eq:max_evidence} yields that
\begin{align}\label{eq:Lgamma}
    \hat{\vgamma}= \mathop{\arg\min}\limits_{\vgamma\in \bR_{+} ^{n}} L(\vgamma): = \vy^{\mathsf{T}}  (\mS(\vgamma) )^{-1}\vy + \log\det  \mS(\vgamma).
\end{align}

Once obtaining the estimate \(\hat{\vgamma} \) of the hyperparameters, we can derive the conditional posterior distribution \(p(\vx|\vy, \hat{\vgamma})\) of the unknown vector \(\vx \) through
\begin{align}\label{eq:cond_posterior}
    p(\vx|\vy,\hat{\vgamma} ) = \frac{p(\vy|\vx)p(\vx|\hat{\vgamma} )}{p(\vy|\hat{\vgamma} )} = \cN(\vx|\vmu(\hat{\vgamma} ), \mSigma(\hat{\vgamma} )),
\end{align}
whose mean and covariance are
\begin{align}\label{eq:mu_Sigma_gamma}
    \mubm(\hat{\vgamma} ) = \hat{\mGamma} \mF^{\mathsf{T}}  (\mS(\hat{\vgamma} ) )^{-1}\vy \quad \mbox{and} \quad
    \mSigma(\hat{\vgamma} ) = \hat{\mGamma}  - \hat{\mGamma} \mF^{\mathsf{T}}  (\mS(\hat{\vgamma} ) )^{-1}\mF\mGamma. 
\end{align}

It is important to highlight that the primary computational expense in the SBL model lies in the calculation of \(\hat{\vgamma} \) \cref{eq:Lgamma}. The associated objective function \(L(\vgamma)\) presents a non-convex nature due to the concavity of the log determinant function, posing significant challenges in solving the minimization problem \cref{eq:Lgamma}. In the following, we review several existing algorithms developed to address this minimization problem. Our focus begins with the widely employed EM algorithm \cite{Dempster1977, wipf2007empirical}. Specifically, the EM algorithm starts with an initial estimate \(\vgamma^{(0)}\), proceeding to iteratively refine the estimates of \(\vgamma\) through a sequence of two key steps:
\begin{itemize}
    \item \textbf{E-step:} Given the current estimate \(\vgamma^{(k)} \), find the posterior distribution \(p(\vx|\vy,\vgamma^{(k)}) \) and then compute the expectation of \(\log p(\vy,\vx|\vgamma)\) with respect to \(p(\vx|\vy,\vgamma^{(k)}) \).
    \item \textbf{M-step:} Update the estimate of \(\vgamma \) by maximizing the expectation obtained in the E-step:
    \begin{align*}
        \vgamma^{(k+1)} = \mathop{\arg\max}\limits_{\vgamma\in \bR_{+} ^{n}}\bE_{\vx|\vy, \vgamma^{(k)}} \log p(\vy,\vx|\vgamma).
    \end{align*}
\end{itemize}
A direct calculation with the conditional posterior distribution \cref{eq:cond_posterior} shows that the EM algorithm updates the estimate of \(\vgamma \) as follows:
\begin{align}\label{eq:EM_scheme}
    \gamma^{k+1}_{i} = [\vmu(\vgamma^{(k)})]_i^2+[\mSigma(\vgamma^{(k)})]_{ii}, \quad 1\leq i\leq n,
\end{align}
where \(\vmu \) and \(\mSigma \) are defined in \cref{eq:mu_Sigma_gamma}. 

The EM algorithm is known to produce a monotonic sequence for the objective function, thus ensuring guaranteed convergence, as demonstrated in \cite{Wu1983, Redner1984}. However, its convergence rate is often slow \cite{Redner1984}, and it exhibits sensitivity to initial values \cite{Ma2005}. In contrast, the MK algorithm \cite{MacKay1992, MacKay1992a, MacKay1999, Tipping2001a} is observed to have significantly faster convergence in many practical applications \cite{Tipping2001a}. The MK algorithm utilizes the following iterative scheme \cite{Tipping2001a}:
\begin{align}\label{eq:MacKay_scheme}
    \gamma_{i} ^{(k+1)} = \gamma_{i} ^{(k)}\frac{[\vmu(\vgamma^{(k)})]_i^2}{\gamma_{i} ^{(k)}-[\mSigma(\vgamma^{(k)})]_{ii}}, \quad 1\leq i\leq n.
\end{align}
However, the MK algorithm currently lacks a theoretical guarantee for convergence.

The CB algorithm, as proposed in \cite{Wipf2007a}, demonstrates a convergence rate comparable to that of the MK algorithm, with the added advantage of guaranteed convergence to a stationary point of the objective function \cite{Wipf2007a}. The CB algorithm employs the following iteration scheme \cite{Wipf2007a} for updating the estimate of \(\vgamma\):
\begin{align}\label{eq:CB_scheme}
    \gamma^{(k+1)}_i  = \gamma_{i} ^{(k)} \sqrt{\frac{[\vmu(\vgamma^{(k)})]_i^2}{\gamma_{i} ^{(k)} - [\mSigma(\vgamma^{(k)})]_{ii}}}, \quad 1\leq i\leq n.
\end{align}

We will present a unified framework for understanding all three algorithms in the next section.

\section{A Unified AML Framework for Hyperparameter Estimation}\label{sec:MM_framework}
We propose a unified framework for the algorithms discussed in \Cref{sec:SBL_intro}: the EM,  MK, and CB algorithms. This framework facilitates their analysis and comparison, and offers a fresh viewpoint for developing more efficient hyperparameter estimation methods in SBL models. Specifically, we introduce an auxiliary variable into the objective function \(L(\boldsymbol{\gamma})\) \cref{eq:Lgamma}, and alternately minimize it over \(\vgamma\) and the auxiliary variable. During the minimization over \(\vgamma\), the non-convex nature of the objective function necessitates the use of a surrogate function to locate the minimizer in each iteration. Through this methodology, it becomes apparent that the EM, MK, and CB algorithms represent distinct approaches of selecting surrogate functions. Notably, all three algorithms utilize linearization techniques to construct these surrogate functions. We designate this approach as the alternating minimization and linearization (AML) framework.

\subsection{A unified AML framework}
We will present a unified AML framework for estimating the hyperparameter \(\hat{\gamma} \) in \cref{eq:Lgamma}. We first rewrite the objective function \cref{eq:Lgamma} by introducing an auxiliary variable. To this end, we define the following function with an auxiliary variable \(\vx\):
\begin{align}\label{eq:Fxgamma}
   F(\vx,\vgamma)  = \beta \|\mF\vx - \vy\|^2  + \vx^{\mathsf T}   \mGamma^{\dagger} \vx + \sum_{i\in I_{\vgamma}}\iota_{\{0\}}(x_i), \quad \vx\in \bR^{n}, \vgamma \in \bR_{+}^{n},
\end{align}
where $\mGamma^\dagger$ is the Moore–Penrose inverse of $\mGamma$ and 
$I_{\vgamma}=\{1\leq i\leq n: \gamma_i=0\}$ records the indices of the zero entries of $\vgamma$. Here  $\iota_A $ the indicator function of a set $A$ is defined as $\iota_A(x)=0$ if $x\in A$ and $\iota_A(x)=\infty$ if $x\notin A$. We point out that $\mGamma^\dagger$ is a diagonal matrix with diagonal entries given by
\begin{align*}
    [\mGamma^\dagger]_{ii} = \begin{dcases}
        \gamma_{i} ^{-1} , &\text{ if }  \gamma_{i} \neq 0;\\
        0, &\text{ if }  \gamma_{i} =0.
    \end{dcases} 
\end{align*}
We next show that the first term in the objective function \(L(\vgamma) \) \cref{eq:Lgamma} can be written as the minimum of \(F(\vx,\vgamma) \) over the auxiliary variable \(\vx \). 

\begin{thm}\label{thm:auxiliary}
For any $\vgamma \in \bR_{+}^{n}$ and $\vy\in \bR^{m}$, the optimization problem 
\begin{equation}\label{model:F}
\min_{\vx\in \bR^{n}} F(\vx,\vgamma)
\end{equation}
has a unique minimizer. Denote this minimizer by $\vx^{*}(\vgamma)$. We have that  
\begin{align*}
\vx^{*}(\vgamma) =  \vmu(\vgamma) \quad \mbox{and} \quad
F(\vx^{*}(\vgamma), \vgamma)=\vy^{\mathsf{T}}  (\mS(\vgamma) )^{-1}\vy,
\end{align*}
where $\vmu(\vgamma)$ is given by \eqref{eq:mu_Sigma_gamma} and $\mS(\vgamma)$ is defined by \eqref{eq:evidence}.
\end{thm}
\begin{proof}
We first prove that the optimization problem has a unique minimizer. To this end, we set \(J_{\vgamma}=\{1\leq i\leq n: \gamma_{i} \neq 0\} \) and  \(I_{\vgamma}=\{1\leq i\leq n: \gamma_{i} = 0\} \), and rewrite \(F(\vx,\vgamma) \) as
\begin{align*}
        F(\vx,\vgamma) = \beta \|\mF\vx - \vy\|^2  + \sum_{i\in J_{\vgamma} } x_{i} ^{2}\gamma_{i}^{-1} + \sum_{i\in I_{\vgamma}}\iota_{\{0\}}(x_i).
\end{align*}
The domain of $F(\cdot,\vgamma)$ is $A=\{\vx: x_i=0, i \in I_{\vgamma}\}$  which is a closed convex set of $\bR^{n}$. On the set $A$,  $F(\cdot,\vgamma)$ is strictly convex and coercive due to the second term $\sum_{i\in J_{\vgamma} } x_{i} ^{2}\gamma_{i}^{-1} $ in $F(\cdot,\vgamma)$, hence, the optimizer of the optimization problem~\eqref{model:F} exists and unique. We denote the minimizer by $\vx^{*}(\vgamma)$. 

Next, we show the minimizer \(\vx^{*}(\vgamma) \) is given by \(\vmu(\vgamma)\). By the definition of \(\vmu(\vgamma)\) in \cref{eq:mu_Sigma_gamma}, we need to show for \(1\leq i\leq n\),
\begin{align}\label{eq:thm:auxiliary:1}
    [\vx^{*}(\vgamma)]_{i} = [\mGamma\mF^{\mathsf{T}} (\mS(\vgamma))^{-1}\vy]_{i}. 
\end{align}
We first show it holds for \(i\in  I_{\vgamma}\). Since  $\vx^{*}(\vgamma) \in A$, the components of the minimizer \([\vx^{*}(\vgamma)]_{i}\) must be zero for \(i\in  I_{\vgamma}\). On the other hand, for \(i\in I_{\vgamma}  \), we have \(\gamma_{i} =0 \) and \([\mGamma\mF^{\mathsf{T}} (\mS(\vgamma))^{-1}\vy]_{i}=0 \), which implies \cref{eq:thm:auxiliary:1} holds for \(i\in I_{\vgamma} \). 

The above equality \cref{eq:thm:auxiliary:1} also holds for \(i\in J_{\vgamma} \). To this end, we use \(F_{J_{\vgamma} }\) to denote the submatrix of columns of \(\mF \) corresponding to \(J_{\vgamma} \), \(\mGamma_{J_{\vgamma}} = \diag(\gamma_{i} : i\in J_{\vgamma} ) \), and \(\vx_{J_{\vgamma} } = [x_{i} ]_{i\in J_{\vgamma} }  \) to denote the corresponding block of \(\vx \). We have 
    \begin{align*}
        [\vx^{*}(\vgamma)]_{J_{\vgamma}} = \mathop{\arg\min}\limits_{\vu\in \bR^{|J_{\vgamma}|}} \beta \|\mF_{J_{\vgamma} }\vu - \vy\|^2  + \vu^{\mathsf{T}}  \mGamma_{J_{\vgamma} }^{-1}\vu.
    \end{align*}
    By setting the gradient to the above quadratic objective function to zero, we have
    \begin{align}\label{eq:thm:auxiliary:2}
        [\vx^{*}(\vgamma)]_{J_{\vgamma}} = (\beta \mF_{J_{\vgamma} }^{\mathsf{T}}  \mF_{J_{\vgamma} } + \mGamma_{J_{\vgamma} }^{-1} )^{-1}\beta \mF_{J_{\vgamma} }^{\mathsf{T}}  \vy.
    \end{align}
    It follows from the Woodbury matrix identity \cite{Higham2002} that 
    \begin{align*}
        (\beta \mF_{J_{\vgamma} }^{\mathsf{T}}  \mF_{J_{\vgamma} } + \mGamma_{J_{\vgamma} }^{-1} )^{-1} = \mGamma_{J_{\vgamma} } - \mGamma_{J_{\vgamma} } \mF_{J_{\vgamma} }^{\mathsf{T}} (\beta^{-1} \mI + \mF_{J_{\vgamma} } \mGamma_{J_{\vgamma} } \mF_{J_{\vgamma} }^{\mathsf{T}} )^{-1} \mF_{J_{\vgamma} } \mGamma_{J_{\vgamma} },
    \end{align*}
    which implies
    \begin{align*}
        [\vx^{*}(\vgamma)]_{J_{\vgamma}} &= \mGamma_{J_{\vgamma} } (\mI - \mF_{J_{\vgamma} }^{\mathsf{T}} (\beta^{-1} \mI + \mF_{J_{\vgamma} } \mGamma_{J_{\vgamma} } \mF_{J_{\vgamma} }^{\mathsf{T}} )^{-1} \mF_{J_{\vgamma} } \mGamma_{J_{\vgamma} }) \beta\mF_{J_{\vgamma} }^{\mathsf{T}} \vy\\
        & = \mGamma_{J_{\vgamma} } \mF_{J_{\vgamma} }^{\mathsf{T}} (\mI - (\beta^{-1} \mI + \mF_{J_{\vgamma} } \mGamma_{J_{\vgamma} } \mF_{J_{\vgamma} }^{\mathsf{T}} )^{-1} \mF_{J_{\vgamma} } \mGamma_{J_{\vgamma} } \mF_{J_{\vgamma} }^{\mathsf{T}}) \beta\vy\\
        & = \mGamma_{J_{\vgamma} } \mF_{J_{\vgamma} }^{\mathsf{T}} (\beta^{-1} \mI + \mF_{J_{\vgamma} } \mGamma_{J_{\vgamma} } \mF_{J_{\vgamma} }^{\mathsf{T}} )^{-1} \vy.
    \end{align*}
    Moreover, since \(\gamma_{i} =0 \) for \(i\in I_{\vgamma}  \), we have \(\mF_{J_{\vgamma} } \mGamma_{J_{\vgamma} } \mF_{J_{\vgamma} }^{\mathsf{T}} = \mF \mGamma \mF^{\mathsf{T}} \). Recalling the definition of \(\mS(\vgamma) \) in \cref{eq:evidence}, we have
    \(
        [\vx^{*}(\vgamma)]_{J_{\vgamma}} = \mGamma_{J_{\vgamma} } \mF_{J_{\vgamma} }^{\mathsf{T}} \mS(\vgamma)^{-1} \vy
    \),
    which implies \cref{eq:thm:auxiliary:1} also holds for \(i\in J_{\vgamma} \). Consequently, \(\vx^{*}(\vgamma) = \mGamma\mF^{\mathsf{T}} (\mS(\vgamma))^{-1}\vy = \vmu(\vgamma)\) is the minimizer of \(F(\vx, \vgamma) \) over \(\vx \) for any \(\vgamma\in \bR^{n}_{+}  \).

    It remains to show the minimum of \(F(\cdot, \vgamma)\) is \(\vy^{\mathsf{T}}  (\mS(\vgamma) )^{-1}\vy \). Since \([\vx^{*}(\vgamma)]_{i}=0 \) for \(i\in I_{\vgamma}  \), 
    \begin{align*}
        F(\vx^{*}(\vgamma), \vgamma) = \beta \|\mF_{J_{\vgamma}} \vx^{*}_{J_{\vgamma} } -\vy\|^2 + (\vx^{*}_{J_{\vgamma} })^{\mathsf{T}}  \mGamma_{J_{\vgamma} }^{-1} \vx^{*}_{J_{\vgamma} }. 
    \end{align*}
    Substituting \(\vx^{*}_{J_{\vgamma} } \) in \cref{eq:thm:auxiliary:2} into the above equation yields that
    \begin{align*}
        F(\vx^{*}_{\vgamma},\vgamma ) = \beta \vy^{\mathsf{T}} (\vy - \mF_{J_{\vgamma}} \vx^{*}_{J_{\vgamma} }) = \vy^{\mathsf{T}}  \left[\beta\mI - \beta^{2} \mF_{J_{\vgamma} } (\beta \mF_{J_{\vgamma} }^{\mathsf{T}}  \mF_{J_{\vgamma} } + \mGamma_{J_{\vgamma} }^{-1} )^{-1} \mF_{J_{\vgamma} }^{\mathsf{T}}\right]\vy.
    \end{align*}
    It follows from the Woodbury matrix identity \cite{Higham2002} that
    \begin{align*}
        (\beta^{-1} \mI + \mF_{J_{\vgamma} }\mGamma_{J_{\vgamma} } \mF_{J_{\vgamma} }^{\mathsf{T}} )^{-1} = \beta\mI - \beta^{2} \mF_{J_{\vgamma} } (\beta \mF_{J_{\vgamma} }^{\mathsf{T}}  \mF_{J_{\vgamma} } + \mGamma_{J_{\vgamma} }^{-1} )^{-1} \mF_{J_{\vgamma} }^{\mathsf{T}},
    \end{align*}
    which implies
    \(
        F(\vx^{*}_{\vgamma},\vgamma ) = \vy^{\mathsf{T}} (\beta^{-1} \mI + \mF_{J_{\vgamma} }\mGamma_{J_{\vgamma} } \mF_{J_{\vgamma} }^{\mathsf{T}} )^{-1} \vy
    \).
    Moreover, since \(\mF_{J_{\vgamma} } \mGamma_{J_{\vgamma} } \mF_{J_{\vgamma} }^{\mathsf{T}} = \mF \mGamma \mF^{\mathsf{T}} \), we have
    \begin{align*}
        F(\vx^{*}_{\vgamma},\vgamma ) = \vy^{\mathsf{T}} (\beta^{-1} \mI + \mF \mGamma \mF^{\mathsf{T}} )^{-1} \vy = \vy^{\mathsf{T}} (\mS(\vgamma) )^{-1} \vy,
    \end{align*}
    which completes the proof.
\end{proof}

We emphasize that the above formulation \cref{eq:Fxgamma} exhibits a slight deviation from the formulation presented in \cite{Wipf2007a} which reads $F_1(\vx,\vgamma)=\beta\|\mF\vx-\vy\|^2+\vx^{\mathsf{T}} \mGamma^{-1}\vx$. We shall notice that the domain of $F_1$ over \(\vgamma \) is $\bR^n_{++}=\{\vu\in \bR^{n}: u_{i} >0, 1\leq i\leq n\} $ which does not consider the situation when some $\gamma_i=0$. It is also ignored in existing literature (e.g. \cite{Wipf2007a, wipf2009unified,hashemi2021unification}). This might bring inconveniences in both the analysis and the computation of the corresponding algorithms. The set \(\bR^{n}_{++} \) is open, which could not ensure the existence of a minimizer of the objective function over \(\vgamma \). It is necessary to analyze this specific situation and develop algorithms that could handle it. In contrast, the proposed formulation \(F(\vx,\vgamma) \) in this paper could handle this situation easily by extending the domain to \(\bR^{n}_{+}  \) as shown in \Cref{thm:auxiliary}. 

We could then rewrite the objective function \(L\) in \eqref{eq:Lgamma} as 
\begin{align}\label{eq:Lgamma-reform}
    L(\vgamma) = F(\vmu(\vgamma), \vgamma) + g(\vgamma) = \mathop{\min}\limits_{\vx\in \bR^{n}} F(\vx,\vgamma) + g(\vgamma),
\end{align} 
where
\begin{align}\label{eq:g}
    g(\vgamma) = \log\det \mS(\vgamma) = \log \det (\beta^{-1}\mI + \mF\mGamma\mF^{\mathsf{T}} ),
\end{align}
and reformulate the minimization problem \cref{eq:Lgamma} through introducing the auxiliary variable \(\vx \):
\begin{align*}
    (\hat{\vgamma} , \hat{\vx}) = \mathop{\arg\min}\limits_{\vgamma\in \bR_{+} ^{n}, \vx\in \bR^{n}} F(\vx,\vgamma) + g(\vgamma).
\end{align*}
A popular approach for solving the above minimization problem involving multiple parameters is the alternating minimization (AM) method, also known as Gauss-Seidel iteration scheme, or block coordinate minimization \cite{Bertsekas1989, Wright2015}. Specifically, for a given initial point \(\vgamma^{(0)} \), each step of the AM method consists of two updates, 
$$
\UOLoverbrace{\vgamma^{(k)} \rightarrow} [\vx^{(k)}]^{\vx-update} \UOLunderbrace{\rightarrow \vgamma^{(k+1)}}_{\vgamma-update},
$$
where 
\begin{align}
\mbox{$\vx$-update:}  \quad  &\vx^{(k)} = \mathop{\arg\min}\limits_{\vx\in \bR^{n}} F(\vx,\vgamma^{(k)}) + g(\vgamma^{(k)})\label{eq:x-update}\\
\mbox{$\vgamma$-update:}  \quad     &\vgamma^{(k+1)} = \mathop{\arg\min}\limits_{\vgamma\in \bR_{+} ^{n}} F(\vx^{(k)},\vgamma) + g(\vgamma). \label{eq:gamma-update}\end{align}

We will investigate the \(\vx\)-update \cref{eq:x-update} and the \(\vgamma \)-update \cref{eq:gamma-update} separately. For the \(\vx\)-update \cref{eq:x-update}, it follows from \Cref{thm:auxiliary} that the minimizer \(\vx^{(k)} \) is given by
\begin{align}\label{eq:xk+1}
    \vx^{(k)} = \vmu(\vgamma^{(k)}) = \mGamma^{(k)}\mF^{\mathsf{T}} (\mS(\vgamma^{(k)}))^{-1}\vy.
\end{align}
It is direct to observe that when \(\gamma_{i}^{(k)} =0 \) for some \(i \), we have \(x^{(k)}_{i}=0 \) for the corresponding $i$. 

For the \(\vgamma \)-update \cref{eq:gamma-update}, we note that when \(x^{(k)}_{i}=0 \) for some \(i \), changing a positive \(\gamma_{i} \) to zero will make both \((\vx^{(k)})^{\mathsf T}   \mGamma^{\dagger} \vx^{(k)} \) and \(\log \det \mS(\vgamma) \) smaller, while \( \sum_{i\in I_{\vgamma}}\iota_{\{0\}}(x^{(k)}_i)\) remains unchanged. It implies \(\gamma^{(k+1)}_{i} =0 \) for such \(i \). On the other hand, when \(x^{(k)}_{i}\neq 0 \) for some \(i \), we must have \(\gamma^{(k+1)}_{i} >0 \). Otherwise, the second term \( \sum_{i\in I_{\vgamma}}\iota_{\{0\}}(x^{(k)}_i)\) will be infinite.

Consequently, when \(\gamma_{i}^{(k)} =0 \) or \(x_{i} ^{(k)}=0 \) happens for some \(i \), both \(x_{i}  \) and \(\gamma_{i}  \) will be zero in all the following iterations. We would then remove the corresponding components from the optimization problem and work on the updates of the other components. Otherwise, we will assume \(\gamma_{i}^{(k)} >0 \) and \(x_{i} ^{(k)}\neq 0 \) for all \(i \) in the following analysis.

We point out that the main challenge of the above optimization problem is the non-convexity of the log determinant function \(g(\vgamma) \). A widely used approach to address this challenge is to derive a surrogate function for the log determinant function. Specifically, at each iteration, we will derive a surrogate \(\tg(\vgamma,\vgamma^{(k)})\) of the log determinant function \(g(\vgamma) \) at the current iterate \(\vgamma^{(k)} \) and then minimize the surrogate function instead to find the next iterate \(\vgamma^{(k+1)} \):
\begin{align}\label{eq:general_surrogate}
    \vgamma^{(k+1)} = \mathop{\arg\min}\limits_{\vgamma\in \bR_{++} ^{n}} F(\vx^{(k)}, \vgamma) + \tg(\vgamma,\vgamma^{(k)}).
\end{align}
The choice of the surrogate function \(\tg\) is crucial for the performance of the algorithm. In particular, it is often chosen as a linear function, which is separable, and the resulting problem~\eqref{eq:general_surrogate} is easy to minimize. In the following, we will demonstrate that different choices of this linearized function within the unified AML framework can give rise to  the EM algorithm \cref{eq:EM_scheme}, the MK algorithm \cref{eq:MacKay_scheme}, and the CB algorithm \cref{eq:CB_scheme}.

In the subsequent subsections, the phrase ``an iterative scheme is equivalent to an algorithm" conveys the idea that, for a given initial estimate, the sequence produced by the iterative scheme is identical to that generated by the algorithm. For nonational simplicity, we use some matlab-like notation. For instance, for a vector $\vs \in \bR^n$, $\vs^{-1}$ denotes elelemnt-by-element reciprocal of $\vs$, and $e^{\vs}$ denotes elelemnt-by-element expential of $\vs$.

\subsection{EM in the unified AML framework}
We will show that the EM algorithm \cref{eq:EM_scheme} can be viewed as the minimizer in the unified AML framework \cref{eq:general_surrogate} with a proper chosen surrogate function \(\tg_{\EM}(\vgamma,\vgamma^{(k)}) \). 

With the help of the matrix determinant lemma \cite{Harville1998} that
\begin{align}\label{eq:determinant_lemma}
    \det (\beta^{-1}\mI + \mF\mGamma\mF^{\mathsf{T}}) = \det (\mGamma^{-1}  + \beta \mF^{\mathsf{T}}\mF)   \cdot \det \mGamma  \cdot \det (\beta^{-1} \mI), 
\end{align}
the log determinant function $g$ can be rewritten as 
\begin{align*}
    g(\vgamma) = \log \det (\beta^{-1}\mI + \mF\mGamma\mF^{\mathsf{T}})  = \log \det (\mGamma^{-1}  + \beta \mF^{\mathsf{T}}\mF)  + \sum_{i=1}^{n}\log \gamma_{i}   -m\log \beta.
\end{align*}
We will derive a surrogate function \(\tg_{\EM}\) of $g$ through approximating the first term \(\log \det (\mGamma^{-1}  + \beta \mF^{\mathsf{T}}\mF) \) of the above identity. This approximation is completed through two steps, namely a change of variable and the first-order Taylor expansion. That is, for $\vgamma \in \bR^{n}_{++}$ setting   
\begin{align*}\label{eq:phi}
 \vs=\vgamma^{-1} \quad \mbox{and} \quad   \phi(\vs) = \log \det (\mS  + \beta \mF^{\mathsf{T}}\mF ),
\end{align*}
where  \(\mS = \diag(\vs) = \mGamma^{-1}\), we derive the surrogate function \(\tg_{\EM}(\vgamma,\vgamma^{(k)}) \) as 
\begin{align*}
    \tg_{\EM}(\vgamma,\vgamma^{(k)}) = \phi(\vs^{(k)}) + \langle\nabla \phi(\vs^{(k)}),\vs - \vs^{(k)} \rangle + \sum_{i=1}^{n}\log \gamma_{i} -m\log \beta,
\end{align*}
where $\vs^{(k)} = (\vgamma^{(k)} )^{-1}$. Note that the log determinant is a concave function \cite{Guler2010}, which implies \(\tg_{\EM}(\vgamma,\vgamma^{(k)}) \) is a majorant of \(g \). The gradient of $\phi$ at $\vs^{(k)}$ is
\begin{align*}
    \nabla \phi (\vs^{(k)}) = \operatorname{diag} ([\mS^{(k)} + \beta \mF^{\mathsf{T}}\mF ]^{-1}), 
\end{align*}
where, by the Woodbury matrix identity \cite{Higham2002},  
\begin{align}\label{eq:woodbury}
    [\mS^{(k)} + \beta \mF^{\mathsf{T}}\mF ]^{-1} = (\mS^{(k)})^{-1}  - (\mS^{(k)})^{-1}  \mF^{\mathsf{T}} (\beta^{-1} \mI + \mF (\mS^{(k)})^{-1}  \mF^{\mathsf{T}} )^{-1} \mF (\mS^{(k)})^{-1} .
\end{align}
From the above identity \eqref{eq:woodbury}, together with  \((\mS^{(k)})^{-1}  = \mGamma^{(k)}\), \(\mS(\vgamma) \) in \cref{eq:evidence} and  \(\mSigma(\vgamma) \) in \cref{eq:mu_Sigma_gamma},  we have
\begin{align*}\label{eq:Sigma_gamma_equi}
    [\mS^{(k)} + \beta \mF^{\mathsf{T}}\mF ]^{-1} = \mSigma(\vgamma^{(k)}),
\end{align*}
which implies
\begin{align}\label{eq:EM_surrogate}
    \tg_{\EM}(\vgamma,\vgamma^{(k)}) = \phi(\vs^{(k)}) + \sum_{i=1}^{n} [\mSigma(\vgamma^{(k)})]_{ii} \left(\frac{1}{\gamma_{i}} - \frac{1}{\gamma^{(k)}_{i} } \right)  + \sum_{i=1}^{n}\log \gamma_{i}   -m\log \beta.
\end{align}

\begin{prop}\label{prop:AM-L-EM}
The iteration scheme
    \begin{align}\label{eq:MM-EM}
        \vgamma^{(k+1)} = \mathop{\arg\min}\limits_{\vgamma\in \bR_{++} ^{n}} F(\vx^{(k)}, \vgamma) + \tg_{\EM}(\vgamma,\vgamma^{(k)}),
    \end{align}
    where \(\tg_{\EM}(\vgamma,\vgamma^{(k)}) \) is the surrogate function in \cref{eq:EM_surrogate}, is equivalent to the EM algorithm \cref{eq:EM_scheme}.
\end{prop}
\begin{proof}
It is direct to observe that both terms in the above objective function \cref{eq:MM-EM} are separable in \(\gamma_{i}  \)'s. That is, we could minimize over \(\gamma_{i}  \) separately for \(1\leq i\leq n \). Specifically, we have
\begin{align*}
    \gamma^{(k+1)}_{i} = \mathop{\arg\min}\limits_{\gamma_{i} >0} \frac{c_{i} }{\gamma_{i} } + \log \gamma_{i},
\end{align*}
where \(c_{i}  =  \left[x_i^{(k)}\right] ^{2} + [\mSigma(\vgamma^{(k)})]_{ii} > 0\). A direct computation gives 
\(
    \gamma^{(k+1)}_{i} = c_{i}  =  \left[x_i^{(k)}\right] ^{2} + [\mSigma(\vgamma^{(k)})]_{ii},
\)
which is the same as the EM algorithm \cref{eq:EM_scheme} given the definition of \(\vx^{(k)} \) in \cref{eq:xk+1}.
\end{proof}

\subsection{MK in the unified AML framework}
We will derive the corresponding surrogate function \(\tg_{\MK}(\vgamma,\vgamma^{(k)}) \) for the MK algorithm \cref{eq:MacKay_scheme} in the unified AML framework. To this end, we consider a change of variable and write the log determinant function in the new variable:
\begin{align*}
    \vgamma=\se^{\vlambda} \quad \mbox{and}\quad   \psi(\vlambda) := g(\se^{\vlambda}). 
\end{align*}
By \cref{eq:determinant_lemma}, we have
\begin{align*}
    \psi(\vlambda) = \log \det (\diag(\se^{-\vlambda})  + \beta \mF^{\mathsf{T}}\mF ) + \sum_{i=1}^{n}\lambda_{i}  -m\log \beta.
\end{align*}

We then use the first-order Taylor expansion of \(\psi(\vlambda) \) at the current iterate \(\vlambda^{(k)} = \log (\vgamma^{(k)}) \) as the surrogate function of $g$:
\begin{align*}
    \tg_{\MK}(\vgamma,\vgamma^{(k)}) = \psi(\vlambda^{(k)}) + \langle\nabla \psi(\vlambda^{(k)}),  \vlambda - \vlambda^{(k)}\rangle.
\end{align*}
By computing the partial derivative \(\frac{\partial \psi}{\partial \lambda_{i} }(\vlambda) = 1 - \se^{-\lambda_{i}} \left[\mGamma^{-1} + \beta\mF^{\mathsf{T}}\mF \right]_{ii}=1 - \gamma_{i}^{-1}  [\mSigma(\vgamma)]_{ii}\), we have
\begin{align}\label{eq:MK_surrogate}
    \tg_{\MK}(\vgamma,\vgamma^{(k)}) = \psi(\vlambda^{(k)}) + \sum_{i=1}^{n} \left(1 - (\gamma_{i} ^{(k)})^{-1} [\mSigma(\vgamma^{(k)})]_{ii}\right) (\log \gamma_{i}  - \log \gamma_{i}^{(k)}). 
\end{align}
We remark that this surrogate function is not guaranteed to be a majorant of \(g(\vgamma) \) since the function \(\psi \) is not concave. 

We next show that the MK algorithm \cref{eq:MacKay_scheme} can be viewed as the minimizer in the unified AML framework \cref{eq:general_surrogate} with the surrogate function \(\tg_{\MK}(\vgamma,\vgamma^{(k)}) \).

\begin{prop}
    The iteration scheme
    \begin{align}\label{eq:MM-MK}
        \vgamma^{(k+1)} = \mathop{\arg\min}\limits_{\vgamma\in \bR_{++} ^{n}} F(\vx^{(k)}, \vgamma) + \tg_{\MK}(\vgamma,\vgamma^{(k)}),
    \end{align}
    where \(\tg_{\MK}(\vgamma,\vgamma^{(k)}) \) is the surrogate function in \cref{eq:MK_surrogate}, is equivalent to the MK algorithm \cref{eq:MacKay_scheme}.
\end{prop}
\begin{proof}
    Since both terms in the above objective function \cref{eq:MM-MK} are separable in \(\gamma_{i}  \)'s, we could minimize over \(\gamma_{i}  \) separately for \(1\leq i\leq n \). Specifically, for \(1\leq i\leq n \), we have
    \begin{align*}
        \gamma^{(k+1)}_{i} &= \mathop{\arg\min}\limits_{\gamma_{i} >0} \frac{(x_i^{(k)})^{2} }{\gamma_{i} } + \left(1 - (\gamma_{i} ^{(k)})^{-1} [\mSigma(\vgamma^{(k)})]_{ii}\right) \log \gamma_{i}
        = \mathop{\arg\min}\limits_{\gamma_{i} >0} \frac{c_{i} }{\gamma_{i} } + \log \gamma_{i},
    \end{align*}
    where \(c_{i} = \frac{(x_i^{(k)})^{2}}{1 - (\gamma_{i} ^{(k)})^{-1} [\mSigma(\vgamma^{(k)})]_{ii}}>0 \). Similar to the proof of \Cref{prop:AM-L-EM}, we have \(\gamma^{(k+1)}_{i} = c_{i}\).
    This combined with the definition of \(\vx^{(k)} \) in \cref{eq:xk+1} implies that the above iteration scheme is equivalent to the MK algorithm \cref{eq:MacKay_scheme}.
\end{proof}

\subsection{CB in the unified AML framework}
The CB algorithm \cref{eq:CB_scheme} can also be considered in  unified AML framework \cref{eq:general_surrogate}. Unlike the EM and MK algorithms, $\tg_{\CB}$ the surrogate function of $g$ is simply the  first-order Taylor expansion of $g$ without any change of variable. That is, 
\begin{align*}
    \tg_{\CB}(\vgamma,\vgamma^{(k)}) = g(\vgamma^{(k)}) + \langle\nabla g(\vgamma^{(k)}),\vgamma - \vgamma^{(k)}\rangle.
\end{align*}

A direct calculation from the definition of \(g(\vgamma) \) in \cref{eq:g} yields that
\begin{align*}
    \nabla g(\vgamma^{(k)}) = \diag(\mF^{\mathsf{T}} (\beta^{-1} \mI + \mF \mGamma^{(k)} \mF^{\mathsf{T}} )\mF).
\end{align*}
From the definition of \(\mSigma(\vgamma) \) in \cref{eq:mu_Sigma_gamma}, we have
\begin{align*}
    \mF^{\mathsf{T}} (\beta^{-1} \mI + \mF \mGamma^{(k)} \mF^{\mathsf{T}} )\mF = (\mGamma^{(k)})^{-1} - (\mGamma^{(k)})^{-1} \mSigma(\vgamma^{(k)}) (\mGamma^{(k)})^{-1},  
\end{align*}
which implies
\(
    \frac{\partial g}{\partial \gamma_{i} } (\vgamma^{(k)}) = (\gamma_{i} ^{(k)})^{-1} - (\gamma_{i} ^{(k)})^{-2} [\mSigma(\vgamma^{(k)})]_{ii}
\).
Consequently, we have
\begin{align}\label{eq:CB_surrogate}
    \tg_{\CB}(\vgamma,\vgamma^{(k)}) = g(\vgamma^{(k)}) + \sum_{i=1}^{n} \left(\gamma_{i} ^{(k)})^{-1} - (\gamma_{i} ^{(k)})^{-2}[\mSigma(\vgamma^{(k)})]_{ii}\right) (\gamma_{i}  - \gamma_{i} ^{(k)}).
\end{align}

We next show that the AML framework \cref{eq:general_surrogate} with the surrogate function \(\tg_{\CB}(\vgamma,\vgamma^{(k)}) \) is equivalent to the CB algorithm \cref{eq:CB_scheme}. 

\begin{prop}
    The iteration scheme
    \begin{align*}\label{eq:MM-CB}
        \vgamma^{(k+1)} = \mathop{\arg\min}\limits_{\vgamma\in \bR_{++} ^{n}} F(\vx^{(k)}, \vgamma) + \tg_{\CB}(\vgamma,\vgamma^{(k)}),
    \end{align*}
    where  \(\tg_{\CB}(\vgamma,\vgamma^{(k)}) \) is the surrogate function in \cref{eq:CB_surrogate}, is equivalent to the CB algorithm \cref{eq:CB_scheme}.
\end{prop}
\begin{proof}
    Similarly, the above objective function is separable in \(\gamma_{i}  \)'s and we could minimize over \(\gamma_{i}  \) separately for \(1\leq i\leq n \):
    \begin{align*}
        \gamma^{(k+1)}_{i} = \mathop{\arg\min}\limits_{\gamma_{i} >0} \frac{(x_i^{(k)})^{2} }{\gamma_{i} } + \left((\gamma_{i} ^{(k)})^{-1} - (\gamma_{i} ^{(k)})^{-2}[\mSigma(\vgamma^{(k)})]_{ii}\right) (\gamma_{i}  - \gamma_{i} ^{(k)}) = \mathop{\arg\min}\limits_{\gamma_{i} >0} \frac{c_{i} }{\gamma_{i} } + \gamma_{i},
    \end{align*}
    where \(c_{i} = \frac{[\vmu(\vgamma^{(k)})]_i^2}{(\gamma_{i} ^{(k)})^{-1} - (\gamma_{i} ^{(k)})^{-2}[\mSigma(\vgamma^{(k)})]_{ii}} \). Consequently, we have \(\gamma^{(k+1)}_{i} = \sqrt{c_{i} }  \), which is exactly the update rule of the CB algorithm \cref{eq:CB_scheme}. 
\end{proof}

We observe that all the three existing algorithms, the EM algorithm \cref{eq:EM_scheme}, the MK algorithm \cref{eq:MacKay_scheme}, and the CB algorithm \cref{eq:CB_scheme}, can be viewed as the minimizer in the unified AML framework \cref{eq:general_surrogate} with different choices of the surrogate function. Specifically, they employ different techniques in approximating the log determinant function \(g(\vgamma) \) in the unified AML framework: the EM algorithm uses a change of variable \(\gamma_{i} = s_{i} ^{-1}  \) and use the first-order Taylor expansion over \(\vs\) as the approximation, the MK algorithm uses a change of variable \(\gamma_{i} = \se^{\lambda_{i} } \) and use the first-order Taylor expansion over \(\vlambda \) as the approximation, and the CB algorithm uses the first-order Taylor expansion over \(\vgamma \) directly.

It remains unclear which choice of the surrogate function is better than the others. In particular, the theoretical convergence of the MK algorithm \cref{eq:MacKay_scheme} is still missing. Both the EM algorithm \cref{eq:EM_scheme} and the CB algorithm \cref{eq:CB_scheme} lie in the class of the majorization-minimization (MM) framework \cite{Hunter2004}, which is known to converge \cite{wipf2009unified,hashemi2021unification}. However, their convergence rates remain undetermined. To address this gap, in \Cref{sec:denoising}, we will investigate the convergence behaviors of these three algorithms and compare their rates in a specialized denoising scenario involving the measurement matrix \(\mF = \mI \). Additionally, we propose a novel algorithm using a different way of linearization and demonstrate its superior convergence rates compared to existing algorithms.

\section{Hyperparameter Estimation for Denoising Problem}\label{sec:denoising}
In this section, we will focus on the specialized setting of the denoising problem. That is, we assume the measurement matrix \(\mF \) in \cref{eq:general_problem} is the identity matrix \(\mI \). This simple setting provides convenience in a theoretical comparison of the existing algorithms including the EM, the MK algorithm, and the CB algorithm. Moreover, we will also propose a novel algorithm under the unified AML framework and show its advantage over the existing algorithms in terms of convergence rates. Its extension to the setting with more general \(\mF \) will be presented in \Cref{sec:general}.

In the denoising context where \(\mF = \mI\), the objective function \(L(\vgamma)\) \cref{eq:Lgamma} becomes separable in terms of each \(\gamma_{i}\). This separability is expressed as follows:
\begin{align*}\label{eq:Lgamma-denoising}
    L(\vgamma) = \sum_{i=1}^{n} y_{i}^{2}(b + \gamma_{i})^{-1} + \sum_{i=1}^{n} \log(b + \gamma_{i}), 
\end{align*}
where \(b\) is defined as \(b = \beta^{-1}\). Furthermore, the three existing algorithms under consideration—namely, the EM algorithm (\cref{eq:EM_scheme}), the MK algorithm (\cref{eq:MacKay_scheme}), and the CB algorithm (\cref{eq:CB_scheme})—exhibit similar separability with respect to each \(\gamma_{i}\). Consequently, our analysis focuses on the convergence of individual \(\gamma_{i}\) components. Specifically, we concentrate on the 1D problem:
\begin{align}\label{eq:1D_problem}
    \min_{\gamma\geq 0} L(\gamma) \coloneqq \frac{y^{2}}{b + \gamma} + \log(b + \gamma).
\end{align}
We see that the minimizer of the above 1D problem \cref{eq:1D_problem} has a closed-form solution:
\begin{align*}
    \gamma^{*} = \max\left\{y^{2} - b, 0 \right\}.
\end{align*}
We will analyze the convergence of the EM algorithm \cref{eq:EM_scheme}, the MK algorithm \cref{eq:MacKay_scheme}, and the CB algorithm \cref{eq:CB_scheme} to the above minimizer \(\gamma^{*}\). To this end, we recall the definition of the \emph{order of convergence} and the \emph{rate of convergence} \cite{Ortega2000}: we say a sequence \(\{\gamma^{(k)} \} \) converges to \(\gamma^{*} \) with order of convergence \(p\) and rate of convergence \(\zeta \) if 
\begin{align*}
    \lim_{k\to \infty} \frac{|\gamma^{(k+1)} - \gamma^{*} |}{|\gamma^{(k)} - \gamma^{*} |^{p}} = \zeta.
\end{align*}

We start with the analysis of the EM algorithm. The EM algorithm \cref{eq:EM_scheme} reduces to the following update rule for the 1D problem \cref{eq:1D_problem}: for an initial value \(\gamma^{(0)} >0 \), 
\begin{align}\label{eq:EM_1D}
    \gamma^{(k+1)} =  y^{2} \left(\frac{\gamma^{(k)}}{b + \gamma^{(k)}} \right)^{2} + \frac{b \gamma^{(k)}}{b + \gamma^{(k)}}, \quad  k\geq 0.
\end{align} 
We have the following result on the convergence of the EM algorithm. 

\begin{prop}\label{prop:EM_1D}
The following statement for the EM algorithm hold:
\begin{enumerate}[(i)]
\item The case of \(y^{2} \geq  b \).  The EM algorithm \cref{eq:EM_1D} converges to \(\gamma^{*} = y^{2} - b \) with order of convergence \(p=1\) and rate of convergence \(\zeta = \frac{b(2y^{2}-b)}{y^{4}}\).

\item  The case of \(y^{2} <  b \). The EM algorithm \cref{eq:EM_1D} converges to \(\gamma^{*} = 0 \) with order of convergence \(p=1\) and rate of convergence \(\zeta = 1 \). Moreover,
    \begin{align*}
        \frac{b}{k+b/\gamma^{(0)}} \leq \gamma^{(k)} \leq \frac{c_{0}}{k+c_{0}/\gamma^{(0)}}, \quad k\geq 0,
    \end{align*}
    where \(c_{0} = y^{2} + b + \frac{b^{2}}{b- y^{2}} \). That is, \(\gamma^{(k)} \) converges to \(\gamma^{*} = 0 \) in the rate of \(O(1/k) \).
\end{enumerate}
\end{prop}
\begin{proof}
    (i) When \(y^{2} \geq  b \), we have \(\gamma^{*} = y^{2} -b \) and by \cref{eq:EM_1D},
    \begin{align*}
        \gamma^{(k+1)} - \gamma^{*} = \frac{b (b+2\gamma^{(k)})}{(b+\gamma^{(k)})^{2}} (\gamma^{(k)} - \gamma^{*}).
    \end{align*}
    Note that \( \frac{b (b+2\gamma^{(k)})}{(b+\gamma^{(k)})^{2}} \in (0,1)\) when \(\gamma^{(k)} > 0 \). Therefore, 
    \(\lim_{k\rightarrow \infty } \gamma^{(k)} = \gamma^{*} \).  Moreover, we have 
    \begin{align*}
        \lim_{k\rightarrow \infty } \frac{|\gamma^{(k+1)} - \gamma^{*} |}{|\gamma^{(k)} - \gamma^{*} |} = \lim_{k\rightarrow \infty } \frac{b (b+2\gamma^{(k)})}{(b+\gamma^{(k)})^{2}} = \frac{b(b+2\gamma^{*})}{(b+\gamma^{*})^{2}} = \frac{b(2y^{2}-b)}{y^{4}}.
    \end{align*}

    (ii) We next consider the case when \(y^{2} < b \). In this case, we have \(\gamma^{*} = 0 \) and 
    \begin{align*}
        \gamma^{(k+1)} - \gamma^{*} = \frac{(y^{2} + b)\gamma^{(k)} + b^{2}}{(b+\gamma^{(k)})^{2}} (\gamma^{(k)} - \gamma^{*}).
    \end{align*}
    Similar arguments as above yield that \(\lim_{k\rightarrow \infty } \gamma^{(k)} = \gamma^{*} = 0 \) and 
    \begin{align*}
        \lim_{k\rightarrow \infty } \frac{|\gamma^{(k+1)} - \gamma^{*} |}{|\gamma^{(k)} - \gamma^{*} |} = \lim_{k\rightarrow \infty } \frac{(y^{2} + b)\gamma^{(k)} + b^{2}}{(b+\gamma^{(k)})^{2}} = 1.
    \end{align*}

    We emphasize that when the rate of convergence is \(1\), we have a sublinear convergence rate, which is slower than any linear convergence rate. We next present a more detailed analysis of how fast it will converge through deriving its lower bound and upper bound. We first show the lower bound of \(\gamma^{(k)} \). It is direct to observe from \cref{eq:EM_1D} that
    \(
        \gamma^{(k+1)} \geq \frac{b \gamma^{(k)}}{b + \gamma^{(k)}},
    \)
    and
    \(
        \frac{1}{\gamma^{(k+1)}} \leq \frac{b + \gamma^{(k)}}{b \gamma^{(k)}} = \frac{1}{b} + \frac{1}{\gamma^{(k)}} 
    \)
    for \(k\geq 0\).
    Thus, we have \(\frac{1}{\gamma^{(k)}} \leq \frac{k}{b} + \frac{1}{\gamma^{(0)}} \), which implies the desired lower bound of \(\gamma^{(k)} \) immediately. 
    
    It remains to show the upper bound of \(\gamma^{(k)} \). It is enough to show 
    \(
        \frac{1}{\gamma^{(k+1)}} \geq \frac{1}{c_{0}} + \frac{1}{\gamma^{(k)}}
    \) for \(k\geq 0\). We have
    \(
         c_{0} = \frac{(y^{2} + b)\gamma^{(k)}}{\gamma^{(k)}} + \frac{b^{2}}{b - y^{2}}\geq \frac{(y^{2} + b)\gamma^{(k)} + b^{2}}{\gamma^{(k)} + b  - y^{2}}.
    \)
    Combined this with the definition of \(\gamma^{(k+1)}\) \cref{eq:EM_1D} yields that \(\frac{1}{c_{0}} + \frac{1}{\gamma^{(k)}} \leq \frac{\gamma^{(k)} + b  - y^{2}}{(y^{2} + b)\gamma^{(k)} + b^{2}} + \frac{1}{\gamma^{(k)}} = \frac{1}{\gamma^{(k+1)}}\), which finishes the proof.
\end{proof}

We continue with the analysis of the MK algorithm. We observe that the MK algorithm \cref{eq:MacKay_scheme} reduces to the following update rule for the 1D problem \cref{eq:1D_problem}: for an initial value \(\gamma^{(0)} >0 \), 
\begin{align}\label{eq:MacKay_1D}
    \gamma^{(k+1)} =  \frac{y^{2}  \gamma^{(k)}}{b + \gamma^{(k)}}, \quad k\geq 0.
\end{align}

We present the following result on the convergence of the MK algorithm.

\begin{prop}\label{prop:MacKay_1D}
The following stamements for the MK algorithm \cref{eq:MacKay_1D} hold: 
\begin{enumerate}[(i)]
\item The case of \(y^{2} > b \).     The MK algorithm \cref{eq:MacKay_1D} converges to \(\gamma^{*} = y^{2} - b \) with order of convergence \(p=1\) and rate of convergence \(\zeta = \frac{b}{y^{2}}\).

\item The case of \(y^{2} <   b \). The MK algorithm \cref{eq:MacKay_1D} converges to \(\gamma^{*} = 0 \) with order of convergence \(p=1\) and rate of convergence \(\zeta = \frac{y^{2}}{b} \). 
    
\item The case of \(y^{2} =  b \). The MK algorithm converges to \(\gamma^{*} = 0 \) in the rate of \(O(1/k) \):
    \begin{align*}
        \gamma^{(k)} = \frac{b}{k+b/\gamma^{(0)}}, \quad k\geq 0.
    \end{align*}
\end{enumerate}
\end{prop}
\begin{proof}
    (i) When \(y^{2} > b \), \(\gamma^{*} = y^{2} -b \). It follows from \cref{eq:MacKay_1D} that
    \begin{align*}
        \gamma^{(k+1)} - \gamma^{*} = \frac{b}{b+\gamma^{(k)}} (\gamma^{(k)} - \gamma^{*}).
    \end{align*}
    Since \(\frac{b}{b+\gamma^{(k)}} \in (0,1) \) when \(\gamma^{(k)} > 0 \), we have \(\lim_{k\rightarrow \infty } \gamma^{(k)} = \gamma^{*} \). Moreover, 
    \begin{align*}
        \lim_{k\rightarrow \infty } \frac{|\gamma^{(k+1)} - \gamma^{*} |}{|\gamma^{(k)} - \gamma^{*} |} = \lim_{k\rightarrow \infty } \frac{b}{b+\gamma^{(k)}} = \frac{b}{y^{2}}.
    \end{align*}

    (ii) When \(y^{2} < b \), we have \(\gamma^{*} = 0 \) and
    \begin{align*}
        \gamma^{(k+1)} - \gamma^{*} = \frac{y^{2}}{b+\gamma^{(k)}} (\gamma^{(k)} - \gamma^{*}).    
    \end{align*}
    Similar arguments could be used to show \(\lim_{k\rightarrow \infty } \gamma^{(k)} = \gamma^{*} \) and 
    \(
        \lim_{k\rightarrow \infty } \frac{|\gamma^{(k+1)} - \gamma^{*} |}{|\gamma^{(k)} - \gamma^{*} |} = \frac{y^{2}}{b}.
    \)

    (iii) In the special case when \(y^{2} = b \), we have \(\gamma^{*} = 0 \) and
    \(
        \frac{1}{\gamma^{(k+1)}} = \frac{1}{\gamma^{(k)}} + \frac{1}{b},
    \)
    which implies \(\frac{1}{\gamma^{(k)}} = \frac{k}{b} + \frac{1}{\gamma^{(0)}} \). The desired result follows immediately.
\end{proof}

We now analyze the convergence of the CB algorithm \cref{eq:CB_scheme}, which reduces to the following update rule for the 1D problem \cref{eq:1D_problem}: for an initial value \(\gamma^{(0)} >0 \),
\begin{align}\label{eq:CB_1D}
    \gamma^{(k+1)} = \gamma^{(k)} \sqrt{\frac{y^{2}}{b + \gamma^{(k)}}}, \quad k\geq 0.
\end{align}

The convergence of the CB algorithm is presented in the following result.

\begin{prop}\label{prop:CB_1D}
The following statements for the CB algorithm \cref{eq:CB_1D} hold:
\begin{enumerate}[(i)]
    \item The case of \(y^{2} > b \). The CB algorithm \cref{eq:CB_1D} converges to \(\gamma^{*} = y^{2} - b \) with order of convergence \(p=1\) and rate of convergence \(\zeta = \frac{b + y^{2}}{2y^{2}}\).

    \item The case of \(y^{2} < b \). The CB algorithm \cref{eq:CB_1D} converges to \(\gamma^{*} = 0 \) with order of convergence \(p=1\) and rate of convergence \(\zeta = \sqrt{\frac{y^{2}}{b}}  \). 
    
    \item The case of \(y^{2} = b \). The CB algorithm converges to \(\gamma^{*} = 0 \) in the rate of \(O(1/k) \):
    \begin{align*}
        \frac{2b}{k + \frac{2b}{\gamma^{(0)}}} \leq \gamma^{(k)} \leq \frac{c_{0}}{k + \frac{c_{0}}{\gamma^{(0)}}}, \quad k\geq 0,
    \end{align*}
    where \(c_{0} = \mathop{\max}\{4b, \sqrt{ 2b\gamma^{(0)} }\} \) and \(\gamma^{(0)} >0 \) is the initial point.
\end{enumerate}
\end{prop}
\begin{proof}
    (i) When \(y^{2} > b \), we have \(\gamma^{*} = y^{2} - b \). We observe that \(h(\gamma) = \gamma \sqrt{\frac{y^{2}}{b + \gamma}} \) is an increasing function on \(\gamma\in (0,\infty ) \). If \(\gamma^{(0)} > \gamma^{*} \), then \(\gamma^{(1)} = h(\gamma^{(0)}) > h(\gamma^{*}) = \gamma^{*} \). By repeating the same argument, we have \(\gamma^{(k)} > \gamma^{*} \) for \(k>0 \). On the other hand, it also implies \(\sqrt{\frac{y^{2}}{b + \gamma^{(k)}}}\in (0,1) \) and \(\gamma^{(k+1)} < \gamma^{(k)} \). Thus, we have the existence of \(\lim_{k\rightarrow \infty } \gamma^{(k)}\) since the sequence \(\{\gamma^{(k)} \} \) is a decreasing sequence with a lower bound \(\gamma^{*} \). Moreover, by taking the limit \(k\to \infty \) on both sides of \cref{eq:CB_1D}, we have \(\lim_{k\rightarrow \infty } \gamma^{(k)} = \gamma^{*} \) and 
    \begin{align*}
        \lim_{k\rightarrow \infty } \frac{|\gamma^{(k+1)} - \gamma^{*} |}{|\gamma^{(k)} - \gamma^{*} |} = \lim_{\gamma^{(k)} \rightarrow \gamma^{*} } \frac{\gamma^{(k)} \sqrt{\frac{y^{2}}{b + \gamma^{(k)}}} - \gamma^{*} }{\gamma^{(k)} - \gamma^{*}} = \frac{b + y^{2}}{2y^{2}}.
    \end{align*}
    
    (ii) When \(y^{2}< b \), we have \(\gamma^{*} = 0 \). Since \(\sqrt{\frac{y^{2}}{b + \gamma^{(k)}}}\in (0,1) \) for \(\gamma^{(k)} > 0 \), we have the sequence \(\{\gamma^{(k)} \} \) is  decreasing and the existence of \(\lim_{k\rightarrow \infty } \gamma^{(k)}\) follows. By taking the limit \(k\to \infty \) on both sides of \cref{eq:CB_1D}, we have \(\lim_{k\rightarrow \infty } \gamma^{(k)} = \gamma^{*} = 0 \) and  
    \begin{align*}
        \lim_{k\rightarrow \infty } \frac{|\gamma^{(k+1)} - \gamma^{*} |}{|\gamma^{(k)} - \gamma^{*} |} = \lim_{\gamma^{(k)} \rightarrow 0 }\sqrt{\frac{y^{2}}{b + \gamma^{(k)}}}  = \sqrt{\frac{y^{2}}{b}}.
    \end{align*}

    (iii) When \(y^{2} = b \), the sequence \(\{\gamma^{(k)} \} \) is decreasing with limit \(\gamma^{*} = 0 \). A direct computation from \cref{eq:CB_1D} yields that
    \(
        \frac{1}{\gamma^{(k+1)}} = \sqrt{ \frac{1}{(\gamma^{(k)})^{2}} + \frac{1}{b \gamma^{(k)}}}.
    \)
    It implies 
    \(
        \frac{1}{\gamma^{(k+1)}} \leq \frac{1}{\gamma^{(k)}} + \frac{1}{2b}
    \) and thus \(\frac{1}{\gamma^{(k)}} \leq \frac{1}{ \gamma^{(0)}} + \frac{k}{2b}\) for \(k\geq 0 \). The desired lower bound follows immediately. To show the upper bound, it is enough to show
    \(
        \frac{1}{\gamma^{(k+1)}} \geq \frac{1}{\gamma^{(k)}} + \frac{1}{c_0}.
    \)
    It is equivalent to \( \left(\frac{1}{\gamma^{(k)}} + \frac{1}{c_0}\right)^{2} \leq \frac{1}{(\gamma^{(k)})^{2}} + \frac{1}{b \gamma^{(k)}}\) or \(\frac{2}{c_{0}\gamma^{(k)}} + \frac{1}{c_{0}^{2}}\leq \frac{1}{b \gamma^{(k)}}\), which is implied by the definition of \(c_{0} \) immediately.
\end{proof}

We observe that the MK algorithm \cref{eq:MacKay_1D} and the CB algorithm \cref{eq:CB_1D} share a similar form in using the factor \(\frac{y^{2}}{b + \gamma^{(k)}}\) and \(\sqrt{\frac{y^{2}}{b + \gamma^{(k)}}} \) respectively to update \(\gamma^{(k+1)}\) from \(\gamma^{(k)} \). Moreover, from \Cref{prop:MacKay_1D} and \Cref{prop:CB_1D}, we observe that the MK algorithm has better convergence rates than the CB algorithm in both cases when \(y^{2} > b \) and \(y^{2} < b \). This motivates us to consider the following iteration scheme:
\begin{align}\label{eq:SQ_1D}
    \gamma^{(k+1)} = \gamma^{(k)} \left(\frac{y^{2}}{b + \gamma^{(k)}} \right)^{2}, \quad k\geq  0.
\end{align}

We will show that this new algorithm can also be reformulated in the unified AML framework \cref{eq:general_surrogate} with a specific choice of the surrogate function. Specifically, we consider the change of variable \(\gamma = \theta^{-2}  \) and use the first-order Taylor expansion over \(\theta \) as the approximation of \(g(\gamma) = \log (b + \gamma) \). That is, we let
\begin{align*}
    \varphi(\theta) := g(\theta^{-2}),
\end{align*}
and define the surrogate function as
\begin{align*}
    \tg_{\SQ}(\gamma,\gamma^{(k)}) = \varphi(\theta^{(k)}) + \varphi^{\prime} (\theta^{(k)}) (\theta - \theta^{(k)}), \quad \theta>0,
\end{align*}
where \(\theta^{(k)} = (\gamma^{(k)})^{-1/2}  \). It is direct to observe that \(\varphi^{\prime} (\theta) = -\frac{2\theta^{-3}}{b + \theta^{-2}} \), which implies
\begin{align}\label{eq:gSQ_1D}
    \tg_{\SQ}(\gamma,\gamma^{(k)}) = \log (b + \gamma^{(k)}) - \frac{2(\gamma^{(k)})^{\frac{3}{2}}}{b + \gamma^{(k)}} (\gamma^{-\frac{1}{2}} - (\gamma^{(k)})^{-\frac{1}{2}}).
\end{align}

The next result shows that the iteration scheme \cref{eq:SQ_1D} is equivalent to the minimizer in the unified AML framework \cref{eq:general_surrogate} with the surrogate function \(\tg_{\SQ}(\gamma,\gamma^{(k)}) \).

\begin{prop}
    The iteration scheme
    \begin{align*}
        \gamma^{(k+1)} = \mathop{\arg\min}\limits_{\gamma\in \bR_{++} } \frac{(x^{(k)}) ^{2}}{\gamma } + \tg_{\SQ}(\gamma,\gamma^{(k)}),
    \end{align*}
    where \(x^{(k)} = \frac{y^{2}\gamma^{(k)}}{b + \gamma^{(k)}} \), is equivalent to the proposed iteration scheme \cref{eq:SQ_1D}.
\end{prop}
\begin{proof}
    Substituting \(\tg_{\SQ}\) in \cref{eq:gSQ_1D} into the above minimization problem yields that
    \begin{align*}
        \gamma^{(k+1)} = \mathop{\arg\min}\limits_{\gamma\in \bR_{++} } \frac{(x^{(k)}) ^{2}}{\gamma } -  \frac{2(\gamma^{(k)})^{\frac{3}{2}}}{b + \gamma^{(k)}}\gamma^{-\frac{1}{2}}.  
    \end{align*}
    By setting the derivative of the above objective function to zero, we have
    \begin{align*}
        \gamma^{(k+1)} = \frac{(x^{(k)})^{4}(b + \gamma^{(k)})^{2}}{(\gamma^{(k)})^{3}} = \gamma^{(k)} \left(\frac{y^{2}}{b + \gamma^{(k)}} \right)^{2},
    \end{align*}
    which is equivalent to the iteration scheme \cref{eq:SQ_1D}.
\end{proof}

We now present the convergence results of the proposed algorithm \cref{eq:SQ_1D}.
\begin{prop}
The following statements for the proposed algorithm \cref{eq:SQ_1D} hold:
\begin{enumerate}[(i)]
\item The case of \(y^{2} > b \). The proposed algorithm \cref{eq:SQ_1D} converges to \(\gamma^{*} = y^{2} - b \) with order of convergence \(p=1\) and rate of convergence \(\zeta = \left\vert \frac{2b}{y^{2}} - 1 \right\vert \).

\item The case of \(y^{2} < b \).  The proposed algorithm \cref{eq:SQ_1D} converges to \(\gamma^{*} = 0 \) with order of convergence \(p=2\) and rate of convergence \(\zeta = \left(\frac{y^{2}}{b} \right)^{2} \). 
    
\item The case of \(y^{2} = b \).   The proposed algorithm converges to \(\gamma^{*} = 0 \) in the rate of \(O(1/k) \):
    \begin{align*}
        \frac{1}{c_{0}k+1/\gamma^{(0)}} \leq \gamma^{(k)} \leq \frac{1}{\frac{2}{b}k+1/\gamma^{(0)}}, \quad k\geq 0,
    \end{align*}
    where \(c_{0} = \frac{2}{b} + \frac{\gamma^{(0)}}{b^{2}} \).
\end{enumerate}
\end{prop}
\begin{proof}
    (i) When \(y^{2} > b \), we have \(\gamma^{*} = y^{2} -b \). We will show convergence of the sequence \(\gamma^{(k)} \) generated by the proposed algorithm \cref{eq:SQ_1D}. It is direct to observe that \(\gamma^{*} \) is a fixed point of the iteration \cref{eq:SQ_1D}. That is,  when \(\gamma^{(k)} = \gamma^{*} \) for some \(k \), we will have \(\gamma^{(n)} = \gamma^{*} \) for all \(n \geq k \) and \(\lim\limits_{k \to \infty} \gamma^{(k)} = \gamma^{*} \) follows immediately. We will assume \(\gamma^{(k)}\neq \gamma^{*} \) for all \(k\). In this case, the sequence \(\{ \gamma^{(k)} \} \) is consisting of two subsequences: one contains all the \(\gamma^{(k)}\)'s greater than \(\gamma^{*}\) and the other contains the all the \(\gamma^{(k)}\)'s less than \(\gamma^{*}\). We will show the convergence of the whole sequence by proving both subsequences are monotone.

    We begin by showing the subsequence of \(\gamma^{(k)}\)'s greater than \(\gamma^{*}\) is monotonically decreasing. For any \(\gamma^{(s)}, \gamma^{(t)} \) adjacent in this subsequence with \(s<t\), we need to show \(\gamma^{(t)} < \gamma^{(s)} \). Since \(\gamma^{(s)} > \gamma^{*} = y^{2}-b\), we observe from the iteration scheme \cref{eq:SQ_1D} that 
    \begin{align*}
        \gamma^{(s+1)} < \gamma^{(s)}.
    \end{align*}
    If \(t=s+1\), then we have \(\gamma^{(t)} < \gamma^{(s)} \) immediately. Otherwise, we have \(\gamma^{(s+1)}, \gamma^{(s+2)}, \ldots, \gamma^{(t-1)} \) are all less than \(\gamma^{*}\). Note that  \(\gamma^{(s+1)} < \gamma^{*} \) implies \(\gamma^{(s+1)} < \gamma^{(s+2)} \) from the iteration scheme \cref{eq:SQ_1D}. By repeating the same argument, we have 
    \begin{align*}
        \gamma^{(s+1)} < \gamma^{(s+2)} < \cdots < \gamma^{(t-1)} < \gamma^{(t)}.
    \end{align*}
    On the other hand, a direct calculation from the iteration scheme \cref{eq:SQ_1D} gives
    \begin{align*}
        \gamma^{(t)} = \frac{y^{4}}{\frac{b^{2}}{\gamma^{(t-1)}} + 2b + \gamma^{(t-1)}}.
    \end{align*}
    Since \(\gamma^{(s+1)} <  \gamma^{(t-1)} < \gamma^{*} = y^{2} -b \), we have
    \begin{align*}
        \gamma^{(t)} < \frac{y^{4}}{\frac{b^{2}}{y^{2}-b} + 2b + \gamma^{(s+1)}} = \frac{y^{4}}{\frac{b^{2}}{y^{2}-b} + 2b + \gamma^{(s)} \left(\frac{y^{2}}{b + \gamma^{(s)}}\right)^{2} },
    \end{align*}
    which implies
    \begin{align*}
        \frac{\gamma^{(t)}}{\gamma^{(s)}} < \frac{y^{4}}{\gamma^{(s)}\left(\frac{b^{2}}{y^{2}-b} + 2b \right) + (\gamma^{(s)})^{2} \left(\frac{y^{2}}{b + \gamma^{(s)}}\right)^{2} }.
    \end{align*}
    We observe that the right hand side of the above inequality is a decreasing function of \(\gamma^{(s)} \). Since \(\gamma^{(s)} > \gamma^{*} = y^{2} -b \), we have \(\frac{\gamma^{(t)}}{\gamma^{(s)}} <1\) by replacing \(\gamma^{(s)} \) with \(y^{2} -b \) in the right hand side of the above inequality. Thus, we have \(\gamma^{(t)} < \gamma^{(s)} \) and the subsequence of \(\gamma^{(k)}\)'s greater than \(\gamma^{*}\) is monotonically decreasing and bounded below by \(\gamma^{*}\). 

    Similarly, we can show the subsequence of \(\gamma^{(k)}\)'s less than \(\gamma^{*}\) is monotonically increasing and bounded above \(\gamma^{*}\). If either one is finite, then the whole sequence \(\{\gamma^{(k)}\}\) must converge and by taking \(k \) to infinity in both sides of the iteration scheme \cref{eq:SQ_1D}, the limit must be the fixed point \(\gamma^{*}\). Otherwise, both subsequences converge and assume their limits are \(\alpha_{1} \geq \gamma^{*} \) and \(\alpha_{2}\leq \gamma^{*}\) respectively. Since both of them are infinite, we could find infinitely many \(\gamma^{(k_{n})}\) such that \(\gamma^{(k_{n})} \) belongs to the first subsequence and \( \gamma^{(k_{n}+1)} \) belongs to the second subsequence for all \(k_{n}\). By taking \(n \) to infinity in both sides of the iteration scheme \cref{eq:SQ_1D} on \(\gamma^{(k_{n})}\) and \(\gamma^{(k_{n}+1)}\), we obtain \(\alpha_{2} = \alpha_{1} \left(\frac{y^{2}}{b + \alpha_{1}}\right)^{2} \geq \alpha_{1}\), which implies \(\alpha_{1} = \alpha_{2} = \gamma^{*}\). Therefore, the limit of the whole sequence \(\{\gamma^{(k)}\}\) exists and equals to \(\gamma^{*}\).

    Moreover, we have
    \begin{align*}
        \lim_{k\rightarrow \infty } \frac{|\gamma^{(k+1)} - \gamma^{*} |}{|\gamma^{(k)} - \gamma^{*} |} = \left\vert\lim_{\gamma^{(k)}  \rightarrow \gamma^{\ast} } \frac{\gamma^{(k)} \left(\frac{y^{2}}{b + \gamma^{(k)}} \right)^{2} - \gamma^{\ast}}{ \gamma^{(k)} - \gamma^{\ast}  } \right\vert =\left\vert \frac{2b}{y^{2}} - 1 \right\vert.
    \end{align*}

    (ii) We next consider the case when \(y^{2} < b \). Similar to the proof of \Cref{prop:CB_1D}, we have the sequence \(\{\gamma^{(k)} \} \) is decreasing and \(\lim_{k\rightarrow \infty } \gamma^{(k)} = \gamma^{*} = 0 \). Moreover,
    \begin{align*}
        \lim_{k\rightarrow \infty } \frac{|\gamma^{(k+1)} - \gamma^{*} |}{|\gamma^{(k)} - \gamma^{*} |} = \lim\limits_{\gamma^{(k)} \to 0} \frac{\gamma^{(k)} \left(\frac{y^{2}}{b + \gamma^{(k)}} \right)^{2}}{\gamma^{(k)}} = \left(\frac{y^{2}}{b} \right)^{2}.
    \end{align*}

    (iii) When \(y^{2} = b \), the sequence \(\{\gamma^{(k)} \} \) is decreasing with limit \(\gamma^{*} = 0 \). Moreover, 
    \begin{align*}
        \frac{1}{\gamma^{(k+1)}} = \frac{1}{\gamma^{(k)}} + \frac{2}{b} + \frac{\gamma^{(k)}}{b^{2}},  \quad k=0,1,\ldots . 
    \end{align*}
    Note that \(0 \leq  \gamma^{(k)} \leq \gamma^{(0)} \) for all \(k\geq 0\). Thus, we have
    \begin{align*}
        \frac{1}{\gamma^{(k)}} + \frac{2}{b} \leq \frac{1}{\gamma^{(k+1)}} \leq \frac{1}{\gamma^{(k)}} + \frac{2}{b} + \frac{\gamma^{(0)}}{b^{2}},
    \end{align*}
    which implies the desired result immediately.
\end{proof}

\begin{figure}[h!]
\centering
\begin{minipage}{0.45\textwidth}
    \centering
    \includegraphics[height=5cm]{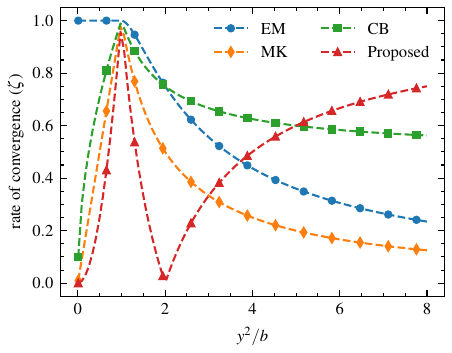}
    \caption{Comparison of convergence rates.}
    \label{fig:convRates1D}
\end{minipage}
\begin{minipage}{0.45\textwidth}
    \centering
    \includegraphics[height=5cm]{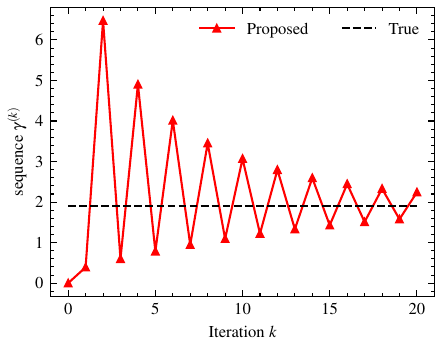}
    \caption{Sequence \(\gamma^{(k)}\) when \(r = \frac{y^{2}}{b} = 20 \).}
    \label{fig:gamma-SQ-1D}
\end{minipage}
\end{figure}

We display in \Cref{fig:convRates1D} the comparison of the convergence rates of the EM algorithm, the MK algorithm, the CB algorithm, and the proposed algorithm. Note that smaller convergence rate \(\zeta \) implies faster convergence. We observe from \Cref{fig:convRates1D} that the proposed algorithm has the best convergence rate when the signal noise ratio \(r = \frac{y^{2}}{b} \) is small (less than 3). However, when \(r\) is large, it is getting worse than the others. This is due to the large deviation of \(\gamma^{(k+1)} \) from \(\gamma^{(k)} \). As displayed in \Cref{fig:gamma-SQ-1D}, \(\gamma^{(k+1)} \) might jump too far away from \(\gamma^{(k)} \) and cause the oscillations around the optimal point. We will discuss how to mitigate this issue in the next section.

\section{An AMQ Hyperparameter Estimation Method}\label{sec:general}
As previously discussed in \Cref{sec:denoising}, the algorithm presented in \cref{eq:SQ_1D} exhibits relatively slow convergence in scenarios where the signal-to-noise ratio \( r = \frac{y^{2}}{b} \) is high. This is primarily attributed to the significant deviation of \(\gamma^{(k+1)}\) from \(\gamma^{(k)}\). In this section, we aim to introduce an AM quadratic (AMQ) method to enhance its convergence for general linear inverse problems \cref{eq:general_problem}. Specifically, after the change of variable \(\vgamma= \vtheta^{-2}\) in \(g(\vgamma)\) (refer to \cref{eq:g}), we will apply the Successive Convex Approximation (SCA) framework to derive the surrogate, following the approach outlined in \cite{scutari2013decomposition}. This involves augmenting the first-order Taylor expansion of \(\vtheta\) with a second-order regularization term. This regularization effectively penalizes the deviation between \(\vtheta^{(k+1)}\) and \(\vtheta^{(k)}\), which, as demonstrated in \cite{scutari2013decomposition}, significantly improves the convergence rate. Additionally, we incorporate a diminishing step size rule within the SCA framework to further optimize the convergence behavior.

We begin by introducing the AMQ method. For a positive constant \(\tau \), we consider the following quadratic surrogate function: 
\begin{align}\label{eq:gSCA}
    \tg_{\SCA}(\vgamma,\vgamma^{(k)}) = \varPsi(\vtheta^{(k)}) + \langle\nabla \varPsi(\vtheta^{(k)}),\vtheta - \vtheta^{(k)}\rangle + \frac{\tau}{2} \|\vtheta - \vtheta^{(k)}\|^2,
\end{align}
where 
\begin{align}\label{eq:varpsi-theta}
    \varPsi(\vtheta) = g(\vtheta^{-2}) = \log \det (\beta\mI + \mF \diag(\vtheta^{-2})\mF^{\mathsf{T}}),  \quad \vtheta \in \bR_{++}^{n}.
\end{align}
We then compute the minimizer of \cref{eq:general_surrogate} with the above surrogate function
\begin{align}\label{eq:theta-k+1/2}
    \vtheta^{(k+\frac{1}{2})} = \mathop{\arg\min}\limits_{\vtheta\in \bR_{++}^{n} } F(\vx^{(k)}, \vgamma) + \tg_{\SCA}(\vgamma,\vgamma^{(k)}),
\end{align}
and update \(\vtheta\) along the direction \(\vtheta^{(k+\frac{1}{2})} - \vtheta^{(k)}\):
\begin{align}\label{eq:theta-k+1}
    \vtheta^{(k+1)} = \vtheta^{(k)} + \eta^{(k)} \left(\vtheta^{(k+\frac{1}{2})} - \vtheta^{(k)} \right),
\end{align}
where \(\eta^{(k)}>0 \) is the step size. Consequently, the corresponding update rule for \(\vgamma\) is given by
\begin{align}\label{eq:AM-SCA}
    \gamma_{i} ^{(k+1)} = (\theta_{i} ^{(k+1)} )^{-2} = \left((\gamma_{i} ^{(k)})^{-\frac{1}{2}} + \eta^{(k)} ((\gamma_{i} ^{(k+\frac{1}{2})})^{-\frac{1}{2}} - (\gamma_{i} ^{(k)})^{-\frac{1}{2}})\right)^{-2}, \quad  1\leq i\leq n,
\end{align}

We point out that the \(\vtheta^{(k+\frac{1}{2})} \) in \cref{eq:theta-k+1/2} could be computed component-wisely since both \(F(\vx, \vgamma) \) and \(\tg_{\SCA}(\vgamma,\vgamma^{(k)}) \) are separable in \(\vtheta \). Specifically, we have
\begin{align*}
    \theta_{i} ^{(k+\frac{1}{2})} = \mathop{\arg\min}\limits_{\theta_{i} >0 } (x_{i}^{(k)})^{2}\theta_{i}^{2}+ [\nabla\varPsi(\vtheta^{(k)})]_{i} \theta_{i} + \tau(\theta_{i} - \theta_{i} ^{(k)})^{2}, \quad 1\leq i\leq n.
\end{align*}
Moreover, it follows from a direct computation from \cref{eq:varpsi-theta} that
\begin{align}\label{eq:nabla-varPsi}
    [\nabla \varPsi(\vtheta)]_{i}  = -2\theta_{i} ^{-3} [Z(\vgamma)]_{ii}, \quad 1\leq i\leq n,
\end{align}
where \(\mZ(\vgamma) = \mF^{\mathsf{T}} (\beta^{-1} \mI + \mF \mGamma\mF^{\mathsf{T}} )^{-1} \mF \). It implies 
\begin{align}\label{eq:gamma-k+1/2}
    \theta_{i} ^{(k+\frac{1}{2})} = \frac{(\theta_{i} ^{(k)})^{-3}\mZ_{ii} ^{(k)} + \tau \theta_{i} ^{(k)}  }{[x_i^{(k)}]^{2} + \tau}, \quad \mbox{and} \quad \gamma_{i} ^{(k+\frac{1}{2})} = \gamma_{i} ^{(k)} \left(\frac{[x_i^{(k)}]^{2} + \tau}{[\gamma_{i} ^{(k)}]^{2} \mZ_{ii}^{(k)} + \tau } \right)^{2}, \quad 1\leq i\leq n.
\end{align}
Here, and in what following, we write \(\mZ^{(k)} = \mZ(\vgamma^{(k)})\) and \(\mZ = \mZ(\vgamma) \) in the simplicity of presentation.

We will next analyze the convergence of the proposed AMQ algorithm \cref{eq:AM-SCA}. To this end, we first derive an upper bound of the Hessian of \(\varPsi(\vtheta)\). The Hessian of \(\varPsi(\vtheta)\) can be directly computed using its gradient in \cref{eq:nabla-varPsi}:
\begin{align*}
    \left[\mH_{\varPsi} (\vtheta)\right]_{i,j} = \frac{\partial^{2} \varPsi(\vtheta)}{\partial \theta_{i} \partial \theta_{j} } = \begin{dcases}
        6\mZ_{ii}\theta_{i} ^{-4} - 4 \mZ_{ii}^{2} \theta_{i} ^{-6}  , &\text{ if } i=j  ;\\
        -4\mZ_{i,j}^{2} \theta_{i} ^{-3}\theta_{j} ^{-3} , &\text{ if } i \neq j.
    \end{dcases}
\end{align*}
We could then use the element-wise (Hadamard) product \(\odot \) to write the Hessian as
\begin{align}\label{eq:Hvarpsi}
    \mH_{\varPsi} (\vtheta) = 6\mZ\odot \mGamma^{2} - 4 \mGamma^{\frac{3}{2}} \odot \mZ \odot \mZ \odot \mGamma^{\frac{3}{2}},
\end{align}
where \(\mGamma = \diag (\theta_{i}^{-2}: 1\leq i\leq n)\). We present its upper bound in the following result.

\begin{lem}\label{lem:Hvarpsi_bound}
   The Hessian \(\mH_{\varPsi} (\vtheta) \)  of \(\varPsi(\vtheta)\) is bounded as
   \begin{align*}
    \mH_{\varPsi} (\vtheta) \preceq \frac{18}{5}\mGamma, \quad \vtheta \in \bR_{++}^{n},
   \end{align*}
   where \(\mA\preceq \mB \) means that \(\mB - \mA\) is positive semidefinite.
\end{lem} 
\begin{proof}
    It follows from a direct computation from \cref{eq:Hvarpsi} that 
    \begin{align}\label{eq:Hessian-varPsi}
        \mH_{\varPsi} (\vtheta) = 6\mGamma^{\frac{3}{2}}\odot \mGamma^{-1} \odot \mZ \odot \mGamma^{-\frac{3}{2}} - 4 \mGamma^{\frac{3}{2}} \odot \mZ \odot \mZ \odot \mGamma^{\frac{3}{2}} = 2 \mGamma^{\frac{3}{2}}\odot(3\mGamma^{-1} -2\mZ) \odot \mZ \odot \mGamma^{\frac{3}{2}}.
    \end{align}
    We will derive the upper bound of \(\mH_{\varPsi}(\vtheta) \) through estimating the bounds of \(\mZ \) and \(3\mGamma^{-1} -2\mZ\). It is direct to observe that \(0 \preceq \mZ\) from the definition of \(\mZ \). On the other hand, by the Woodbury matrix identity \cite{Higham2002}, we have
    \begin{align*}
        (\mGamma^{-1} + \beta\mF^{\mathsf{T}}\mF)^{-1} = \mGamma - \mGamma \mF^{\mathsf{T}} (\beta^{-1}\mI + \mF\mGamma\mF^{\mathsf{T}} )^{-1} \mF \mGamma = \mGamma - \mGamma\mZ\mGamma,
    \end{align*}
    which implies
    \begin{align*}
        \mZ = \mGamma^{-1} - \mGamma^{-1} (\mGamma^{-1} + \beta\mF^{\mathsf{T}}\mF)^{-1}\mGamma^{-1}.
    \end{align*}
    Thus, we have \(0 \preceq \mZ \preceq \mGamma^{-1} \) and \(0 \preceq 3\mGamma^{-1} -2\mZ \preceq 3\mGamma^{-1} \). By the Schur Product Theorem \cite{Bapat1997}, the element-wise product of positive semidefinite matrices is also positive semidefinite. That is, if \(\mA\preceq \mB\) and \(0\preceq \mC\), then \(\mA\odot \mC \preceq \mB\odot \mC \). It follows from substituting \(\mZ \preceq \mGamma^{-1}  \) into \cref{eq:Hessian-varPsi} that 
    \begin{align*}
        \mH_{\varPsi} (\vtheta) \preceq 2 \mGamma^{\frac{3}{2}}\odot(3\mGamma^{-1} -2\mZ) \odot \mGamma^{-1}  \odot \mGamma^{\frac{3}{2}} = 2 \mGamma^{2}\odot(3\mGamma^{-1} -2\mZ)
    \end{align*}
    On the other hand, substituting \(3\mGamma^{-1} -2\mZ \preceq 3\mGamma^{-1} \) into \cref{eq:Hessian-varPsi} gives
    \begin{align*}
        \mH_{\varPsi} (\vtheta) \preceq 2 \mGamma^{\frac{3}{2}} \odot 3\mGamma^{-1} \odot \mZ \odot \mGamma^{\frac{3}{2}} = 6\mGamma^{2}\odot \mZ. 
    \end{align*}
    Consequently, we have
    \begin{align*}
        \mH_{\varPsi} (\vtheta) \preceq \frac{3}{5} [2 \mGamma^{2}\odot(3\mGamma^{-1} -2\mZ)] + \frac{2}{5} [6\mGamma^{2}\odot \mZ] = \frac{18}{5}\mGamma.
    \end{align*}

\end{proof}

\begin{prop}\label{prop:Lgamma-diff}
    Let \(\{\vgamma^{(k)}\}_{k\in \bN} \) be the sequence generated from the proposed AMQ algorithm \cref{eq:AM-SCA} with an initial point \(\vgamma^{(0)}\in \bR_{++}^{n} \). If we choose \(\eta^{(k)} \) such that \(L(\vgamma^{(k+1)})\leq  L(\vgamma^{(k)}) \) for every \(k\geq 0 \), then there exists a positive constant \(R \) such that \(\gamma_{i} ^{(k)}\leq  R \) for all \(1\leq i\leq n \) and \(k\geq 0 \), and \(\varPsi(\vtheta) \) is \(\frac{18R}{5} \)-smooth when \(\theta_{i} \geq R^{-\frac{1}{2}} \) for all \(1\leq i\leq n \). Moreover, we have
    \begin{align*}
        L(\vgamma^{(k+1)}) \leq & L(\vgamma^{(k)}) - \left(\beta \sigma_{\min}^2(\mF) + \frac{1}{R}\right)\left\lVert \vx^{(k+1)} - \vx^{(k)} \right\rVert^{2} - A_{k} (\eta^{(k)}) + B_{k} (\eta^{(k)}), 
    \end{align*}
    where \(\sigma_{\min}(\mF) \) is the smallest singular value of \(\mF \), 
    \(
        A_{k} (\eta^{(k)}) = \eta^{(k)} \sum\limits_{i=1}^{n} \left[(x_i^{(k)})^{2} + \tau\right] (\theta_{i} ^{(k+\frac{1}{2})} - \theta_{i} ^{(k)})^{2}
    \)
    and
    \(
        B_{k}(\eta^{(k)}) = \left(\eta^{(k)} \right)^{2}\sum\limits_{i=1}^{n} \left[(x_i^{(k)})^{2} + \frac{9R}{5}  \right] (\theta_{i} ^{(k+\frac{1}{2})} - \theta_{i} ^{(k)})^{2}.
    \)
\end{prop}
\begin{proof}
    We first show all \(\gamma_i^{(k)}\) are bounded. By the definition of \(L(\vgamma) \) in \cref{eq:Lgamma}, we have
    \begin{align*} 
        \log \det (\beta^{-1} \mI + \mF\mGamma^{(k)}\mF^{\mathsf{T}})  \leq   L(\vgamma^{(k)}) \leq L(\vgamma^{(0)}), \quad  k\geq 0.
    \end{align*}
    Moreover, we observe that \( \det (\beta^{-1} \mI + \mF\mGamma\mF^{\mathsf{T}} )  = \det (\beta^{-1}\mI + \sum_{i=1}^{n} \gamma_{i} F_{i} F_{i} ^{\mathsf{T}})  \) is a monotonically increasing function on each component \(\gamma_i\) and it goes to infinity if any \(\gamma_i\) goes to infinity. Since it is also continuous, there exists a \(R>0\) such that \(\gamma_{i}^{(k)}\leq R\) for all \(1\leq i\leq n\) and \(k\geq 0\). On the other hand, when \(\theta_{i} \geq R^{-\frac{1}{2}} \) for all \(1\leq i\leq n \), we have \(\gamma_{i} \leq R \) and by \Cref{lem:Hvarpsi_bound}, \(\mH_{\varPsi}(\vtheta)  \preceq \frac{18R}{5} \mI \). It implies \(\varPsi(\vtheta)\) is \(\frac{18R}{5}\)-smooth.

    We continue to estimate the difference between \(L(\vgamma^{(k+1)})\) and \(L(\vgamma^{(k)})\). From the reformulation of \(L(\vgamma) \) in \cref{eq:Lgamma-reform}, the definition of \(\vx^{(k)} \) in \cref{eq:xk+1}, and the definition of \(\varPsi(\vtheta) \) in \cref{eq:varpsi-theta}, we have
    \begin{align*}
        L(\vgamma^{(k+1)}) - L(\vgamma^{(k)}) = F(\vx^{(k+1)}, \vgamma^{(k+1)}) - F(\vx^{(k)}, \vgamma^{(k)}) + \varPsi(\vtheta^{(k+1)}) - \varPsi(\vtheta^{(k)}).
    \end{align*}
    We will estimate the two differences of \(F\) and \(\varPsi\) separately. 

    We first estimate the difference of \(F\). 
    By the strong convexity of \(F(\vx, \vgamma^{(k+1)})\) with respect to \(\vx\) and the definition of \(\vx^{(k+1)}\) being the minimizer of \(F(\cdot, \vgamma^{(k+1)})\), we have
    \begin{align*}
        F(\vx^{(k+1)}, \vgamma^{(k+1)}) - F(\vx^{(k)}, \vgamma^{(k+1)}) \leq  -\left(\beta \sigma_{\min}^2(\mF) + \frac{1}{R}\right) \left\lVert \vx^{(k+1)} - \vx^{(k)} \right\rVert^{2}.
    \end{align*}
    On the other hand, a direct computation yields
    \begin{align*}
        F(\vx^{(k)}, \vgamma^{(k+1)}) - F(\vx^{(k)}, \vgamma^{(k)}) = \sum_{i=1}^{n} \left(x_i^{(k)}\right)^{2}\left((\theta_{i} ^{(k+1)})^{2} - (\theta_{i} ^{(k)})^{2} \right).
    \end{align*}
    Thus, we have
    \begin{align*}
        F(\vx^{(k+1)}, \vgamma^{(k+1)}) - F(\vx^{(k)}, \vgamma^{(k)}) \leq & -\left(\beta \sigma_{\min}^2(\mF) + \frac{1}{R}\right) \left\lVert \vx^{(k+1)} - \vx^{(k)} \right\rVert^{2} + \\
        &\sum_{i=1}^{n} \left(x_i^{(k)}\right)^{2}\left((\theta_{i} ^{(k+1)})^{2} - (\theta_{i} ^{(k)})^{2} \right).
    \end{align*}

    We next estimate the difference of \(\varPsi\). Since \(\varPsi\) is \(\frac{18}{5}R\)-smooth, we have
    \begin{align*}
        \varPsi(\vtheta^{(k+1)}) - \varPsi(\vtheta^{(k)}) &\leq \langle\nabla \varPsi(\vtheta^{(k)}),\vtheta^{(k+1)} - \vtheta^{(k)}\rangle + \frac{9R}{5}\lVert\vtheta^{(k+1)} - \vtheta^{(k)}\rVert^{2}\\
        &= \eta^{(k)} \langle\nabla \varPsi(\vtheta^{(k)}),\vtheta^{(k+\frac{1}{2})} - \vtheta^{(k)}\rangle + \frac{9R}{5}(\eta^{(k)})^{2}\lVert\vtheta^{(k+\frac{1}{2})} - \vtheta^{(k)}\rVert^{2}.
    \end{align*}
    We note from the definition of \(\tg_{\SCA}  \) in \cref{eq:gSCA} that 
    \begin{align*}
        \tg_{\SCA} (\vgamma^{(k+\frac{1}{2})}, \vgamma^{(k)}) - \tg_{\SCA}(\vgamma^{(k+\frac{1}{2})}, \vgamma^{(k)}) = \langle\nabla \varPsi(\vtheta^{(k)}), \vtheta^{(k+\frac{1}{2})} - \vtheta^{(k)}\rangle + \tau \left\lVert\vtheta^{(k+\frac{1}{2})} - \vtheta^{(k)} \right\rVert ^{2},
    \end{align*}
    which implies
    \begin{align*}
        \varPsi(\vtheta^{(k+1)}) - \varPsi(\vtheta^{(k)}) \leq &\eta^{(k)} (\tg_{\SCA}(\vgamma^{(k+\frac{1}{2})}, \vgamma^{(k)}) - \tg_{\SCA}(\vgamma^{(k)}, \vgamma^{(k)})) - \eta^{(k)} \tau \left\lVert\vtheta^{(k+\frac{1}{2})} - \vtheta^{(k)} \right\rVert ^{2} + \\
        &\frac{9R}{5}(\eta^{(k)})^{2}\lVert\vtheta^{(k+\frac{1}{2})} - \vtheta^{(k)}\rVert^{2}.
    \end{align*}
    Moreover, by the definition of \(\theta_i^{(k+\frac{1}{2})}\) in \cref{eq:theta-k+1/2}, we have
    \begin{align*}
        \tg_{\SCA}(\vgamma^{(k+\frac{1}{2})}, \vgamma^{(k)}) - \tg_{\SCA}(\vgamma^{(k)}, \vgamma^{(k)}) \leq \sum_{i=1}^{n} \left[(x_i^{(k)})^{2} \right] \left((\theta_{i} ^{(k)})^{2} - (\theta_{i} ^{(k+\frac{1}{2})})^{2} \right).
    \end{align*}
    It follows that
    \begin{align*}
        \varPsi(\vtheta^{(k+1)}) - \varPsi(\vtheta^{(k)}) \leq &\eta^{(k)} \sum_{i=1}^{n} \left[(x_i^{(k)})^{2} \right]  \left((\theta_{i} ^{(k)})^{2} - (\theta_{i} ^{(k+\frac{1}{2})})^{2} \right)  - \eta^{(k)} \tau \left\lVert\vtheta^{(k+\frac{1}{2})} - \vtheta^{(k)} \right\rVert ^{2} + \\
        &\frac{9R}{5}(\eta^{(k)})^{2}\lVert\vtheta^{(k+\frac{1}{2})} - \vtheta^{(k)}\rVert^{2}.
    \end{align*}

    Combining the above estimates on the differences of \(F\) and \(\varPsi\), we obtain
    \begin{align*}
        L(\vgamma^{(k+1)}) - L(\vgamma^{(k)}) \leq &-\left(\beta \sigma_{\min}^2(\mF) + \frac{1}{R}\right) \left\lVert \vx^{(k+1)} - \vx^{(k)} \right\rVert^{2} \\
        & + \sum_{i=1}^{n} \left(x_i^{(k)}\right)^{2}\left((\theta_{i} ^{(k+1)})^{2} - (\theta_{i} ^{(k)})^{2} + \eta^{(k)}\left((\theta_{i} ^{(k)})^{2} - (\theta_{i} ^{(k+\frac{1}{2})})^{2} \right)  \right)\\
        & - \eta^{(k)} \tau \left\lVert\vtheta^{(k+\frac{1}{2})} - \vtheta^{(k)} \right\rVert ^{2} + \frac{9R}{5}(\eta^{(k)})^{2}\lVert\vtheta^{(k+\frac{1}{2})} - \vtheta^{(k)}\rVert^{2}.
    \end{align*}
    By plugging the definition of \(\vtheta^{(k+1)}\) \cref{eq:theta-k+1}, we have
    \begin{align*}
        (\theta_{i} ^{(k+1)})^{2} - (\theta_{i} ^{(k)})^{2} + \eta^{(k)}\left((\theta_{i} ^{(k)})^{2} - (\theta_{i} ^{(k+\frac{1}{2})})^{2} \right) = \left(- \eta^{(k)} + (\eta^{(k)})^{2} \right) \left(\theta_{i} ^{(k+\frac{1}{2})} - \theta_{i} ^{(k)}\right)^{2},
    \end{align*}
    which implies the desired result immediately.
\end{proof}

We are now ready to present the convergence of the sequence of \(\{\vgamma^{(k)}\}_{k\in \bN}\) generated by the proposed algorithm \cref{eq:AM-SCA}.

\begin{thm}
    If we choose \(\eta^{(k)} \) such that the constants \(A_k(\vgamma^{(k)}) \) and \(B_k(\vgamma^{(k)}) \) in \Cref{prop:Lgamma-diff} satisfies \(B_k(\vgamma^{(k)}) \leq A_k(\vgamma^{(k)})\) for all \(k\geq 0 \), where the sequence \(\{\vgamma^{(k)}\}_{k\in \bN} \) is generated by the proposed algorithm \cref{eq:AM-SCA} with an initial point \(\vgamma^{(0)}\in \bR_{++}^{n} \), then we have the following convergence results:
    \begin{enumerate}[(i)]
        \item The sequence \(\{L(\vgamma^{(k)})\}\) is monotonically decreasing and converges.
        \item There exists a subsequence \(\{n_{k}\}\) of \(\bN\) such that both \(\{\vx^{(n_{k})}\} \) and \(\{\vgamma^{(n_{k})}\} \) converge.
        \item For the subsequence \(\{n_{k}\}\) in (ii), if additionally \(\sum_{k=0}^{\infty}\eta^{(n_{k})} = \infty \) and there exists a constant \(\kappa\in [0,1) \) such that \( B_{k}(\vgamma^{(k)}) \leq \kappa A_{k}(\vgamma^{(k)})\) for all \(k\in \bN \), then for each \(1\leq i\leq n \), either \(\gamma_{i}^{*} =0 \) or \(\frac{\partial L(\vgamma^{*})}{\partial \gamma_{i} }=0  \) where \(\vgamma^{*} = \lim_{k\to \infty} \vgamma^{(n_{k})} \).
    \end{enumerate}

\end{thm}
\begin{proof}
    (i) It is direct to observe from \Cref{prop:Lgamma-diff} that when \(A_{k} (\eta^{(k)})\leq 0 \), the sequence \(\{L(\vgamma^{(k)})\}\) is monotonically decreasing. Moreover, from the definition of \(L(\vgamma) \) in \cref{eq:Lgamma}, we know it is always bounded below by \(\log \left\vert \beta^{-1} \mI \right\vert  \), which implies the sequence \(\{L(\vgamma^{(k)})\}\) converges. 
            
    (ii) By \Cref{prop:Lgamma-diff}, we have \(\{\vgamma^{(k)}\}\) is bounded. Thus, there exists a subsequence \(\{n_{k}\}\) of \(\bN\) such that \(\{\vgamma^{(n_{k})}\} \) converges. By the definition of \(\vx^{(k)} \) in \cref{eq:xk+1}, we have \(\{\vx^{(n_{k})}\} \) converges as well.

    (iii)  We will prove the desired result on \(\vgamma^{*}\) through first showing \(\lim_{k\to \infty}\vgamma^{(n_{k}+\frac{1}{2})} =\vgamma^{*}\) as well and then taking the limit on both sides of \cref{eq:gamma-k+1/2}. Since \( B_k(\vgamma^{(k)}) \leq \kappa A_k(\vgamma^{(k)})\) for all \(k\in \bN \), by \Cref{prop:Lgamma-diff} we have \(L(\vgamma^{(k+1)}) - L(\vgamma^{(k)}) \leq - (1-\kappa) A_{k} (\eta^{(k)}) \). It follows from the convergence of \({L(\vgamma^{(k)})} \) that \(\sum_{k=0}^{\infty} A_{k} (\vgamma^{(k)})<\infty \), which implies \(\sum_{k=0}^{\infty} \eta^{(k)} \tau \left\lVert \vtheta^{(k+\frac{1}{2})} - \vtheta^{(k)}\right\rVert ^{2}<\infty  \). Moreover, since \(\{\vgamma^{(n_{k})}\} \) converges, we have the convergence of \(\{\vtheta^{(n_{k})}\} \). The convergence of \(\{\vtheta^{(n_{k}+\frac{1}{2})}\} \) follows immediately from \cref{eq:theta-k+1/2}. This implies \(\lim_{k\to \infty} \left\lVert \vtheta^{(k+\frac{1}{2})} - \vtheta^{(k)}\right\rVert\) exists. Since \(\sum_{k=0}^{\infty}\eta^{(n_{k})} = \infty \), we have \(\lim_{k\to \infty} \left\lVert \vtheta^{(n_{k}+\frac{1}{2})} - \vtheta^{(n_{k})}\right\rVert =0\) and \(\lim_{k\to \infty}\vtheta^{(n_{k}+\frac{1}{2})} = \lim_{k\to \infty}\vtheta^{(n_{k})} \). That is, \(\lim_{k\to \infty}\vgamma^{(n_{k}+\frac{1}{2})} = \lim_{k\to \infty}\vgamma^{(n_{k})} =\vgamma^{*}\). 
            
    Taking the limit on both sides of the definition of \(\vgamma^{(n_{k}+\frac{1}{2})} \) in \cref{eq:gamma-k+1/2}, we have
    \begin{align*}
        \gamma_{i} ^{*} = \gamma_{i} ^{*} \left(\frac{[\vx^{*}(\vgamma^{*})]_{i} ^{2} + \tau}{(\gamma_{i} ^{*})^{2} [Z(\vgamma^{*})]_{ii} + \tau}\right)^{2}, \quad 1\leq i\leq n.
    \end{align*}
    It implies either \(\gamma_{i} ^{*}=0 \) or \([\vx^{*}(\vgamma^{*})]_{i} ^{2} = (\gamma_{i} ^{*})^{2} [Z(\vgamma^{*})]_{ii} \). When \(\gamma_{i} ^{*}\neq 0 \), we have from the formula of \(\vx^{(k)} \) \cref{eq:xk+1} that \([\vu(\vgamma^{*})]_{i} ^{2} = [Z(\vgamma^{*})]_{ii} \), where \(\vu(\vgamma) = \mF^{\mathsf{T}} (\mS(\vgamma))^{-1} \vy \). On the other hand, a direct calculation from \(L(\vgamma) \) \cref{eq:Lgamma} yields \(\frac{\partial L(\vgamma)}{\partial \gamma_{i} } = [Z(\vgamma)]_{ii} - [\vu(\vgamma)]_{i} ^{2} \). Consequently, for each \(1\leq i\leq n \), we have either \(\gamma_{i} ^{*}=0 \) or \(\frac{\partial L(\vgamma^{*})}{\partial \gamma_{i} }=0 \).    
\end{proof}

It should be noted that the condition \( B_k(\vgamma^{(k)}) \leq \kappa A_k(\vgamma^{(k)}) \) for all \( k \in \bN \) is sufficient, but not necessary, for the convergence of the sequence \(\{\vgamma^{(k)}\}\). While this condition offers a conservative approach for selecting the step size \(\eta^{(k)}\), it is often impractical to use it directly due to the challenge in estimating the upper bound \(R\). Therefore, in our numerical experiments, we will adopt a more pragmatic approach for choosing the step size \(\eta^{(k)}\), as suggested in \cite{scutari2013decomposition}:
\begin{align}\label{eq:eta-k}
    \eta^{(k)} = \eta^{(k-1)} (1 - \epsilon \eta^{(k-1)}), \quad \text{for some } \epsilon \in (0,1).
\end{align}
It provides a diminishing step size with \(\eta^{(k)} \to 0 \) as \(k\to \infty \). It is easy to observe that \(B_k(\vgamma^{(k)}) \leq \kappa A_k(\vgamma^{(k)})\) is satisfied when \(\eta^{(k)}\leq \frac{5\kappa\tau}{9R}\), which is guaranteed for large enough \(k\).

\section{Numerical Experiments}\label{sec:numerical}
We will present several numerical experiments to demonstrate the performance of our proposed algorithm AMQ \cref{eq:AM-SCA} for the general linear inverse problem \cref{eq:general_problem}. We will compare it with the EM algorithm \cref{eq:EM_scheme}, the MK algorithm \cref{eq:MacKay_scheme}, and the CB algorithm \cref{eq:CB_scheme}. We will consider both synthetic data and real data in the experiments. 

\subsection{Synthetic data for linear inverse problems}
We first consider the general linear inverse problem \cref{eq:general_problem} with synthetic data. We will test the performance of the proposed algorithm with different matrices \(\mF\). Specifically, we will consider two cases: the identity matrix for the denoising problem and the partial DCT matrix for the Fourier reconstruction problem. 

For both cases, we generate the true signal \(\vx\) as a sparse vector with \(s\%\) non-zero entries. We will then generate the measurement \(\vy\) according to \cref{eq:general_problem} with a given noise level \(\beta^{-1}\). We will test the performance of the proposed algorithm with different noise levels $\{ 10^{-1}, 1, 10\}$ and with different sparsity levels (\(s=10, 80\)). In all the experiments, the regularization parameter \(\tau\) is set to be \(10^{-10}\) and the step size \(\eta^{(k)}\) is generated by the formula in \cref{eq:eta-k} with \(\epsilon=0.02\) and \(\eta^{(0)}=1\). We will compare the performance of the proposed algorithm with the EM algorithm \cref{eq:EM_scheme}, the MK algorithm \cref{eq:MacKay_scheme}, and the CB algorithm \cref{eq:CB_scheme}. For all the algorithms, we will use the same initial point \(\vgamma^{(0)}\) and the stopping criterion when the relative change of \(\vgamma^{(k)}\) is less than \(10^{-3}\).

\begin{figure}[h!]
    \centering
    \begin{subfigure}[b]{0.32\textwidth}
        \includegraphics[height=3.5cm]{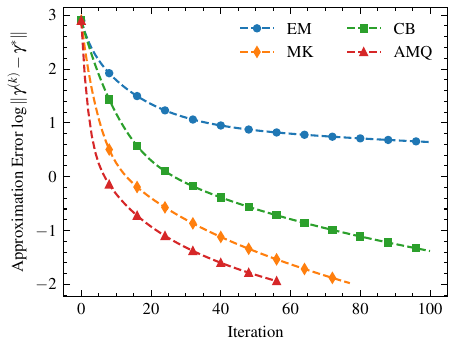}
    \end{subfigure}
    \begin{subfigure}[b]{0.32\textwidth}
        \includegraphics[height=3.5cm]{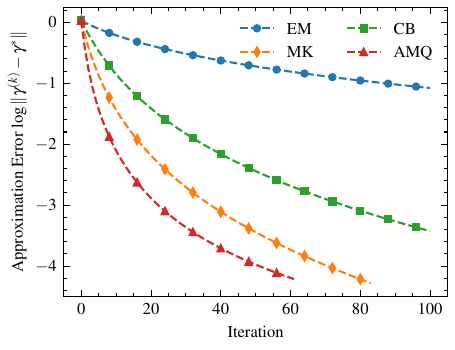}
    \end{subfigure}
    \begin{subfigure}[b]{0.32\textwidth}
        \includegraphics[height=3.5cm]{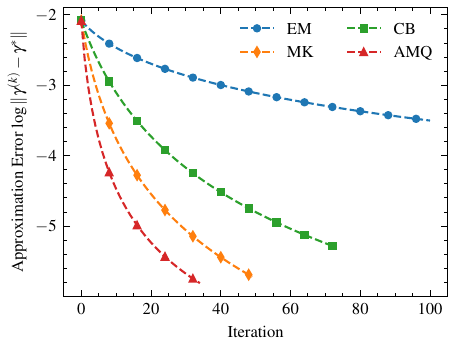}
    \end{subfigure}

    \begin{subfigure}[b]{0.32\textwidth}
        \includegraphics[height=3.5cm]{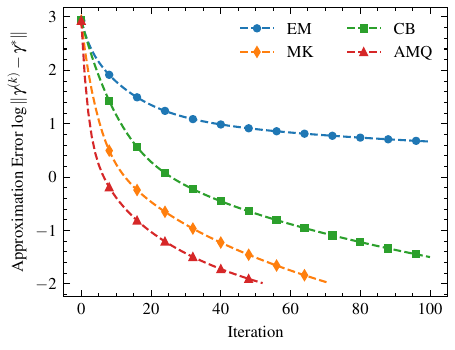}
    \end{subfigure}
    \begin{subfigure}[b]{0.32\textwidth}
        \includegraphics[height=3.5cm]{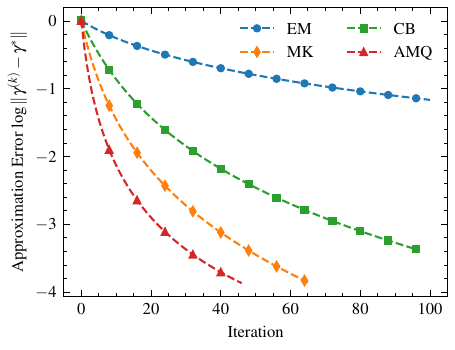}
    \end{subfigure}
    \begin{subfigure}[b]{0.32\textwidth}
        \includegraphics[height=3.5cm]{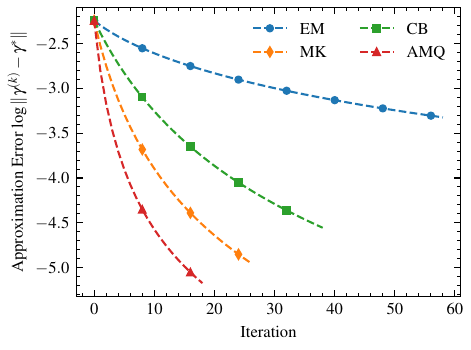}
    \end{subfigure}

    \caption{Approximation errors for the denoising problem. Top row: \(s=10\), bottom row: \(s=80\). Left column: \(\beta=10^{-1}\), middle column: \(\beta=1\), right column: \(\beta=10\). }
    \label{fig:identity-gamma}
\end{figure}

For the denoising case, the matrix \(\mF\) is set to be the \(512\times 512\) identity matrix. The optimal solution \(\vgamma^{*}\) has a closed form \(\gamma_{i}^{*} = \max\{0, y_{i}^{2}-\beta^{-1}\} \) for \(1\leq i\leq n\). We display the logorithm of the approximation errors \(\log\left\lVert \vgamma^{(k)} - \vgamma^{*}\right\rVert \) in \Cref{fig:identity-gamma} for different noise levels and sparsity levels.

\begin{figure}[h!]
    \centering
    \begin{subfigure}[b]{0.32\textwidth}
        \includegraphics[height=3.5cm]{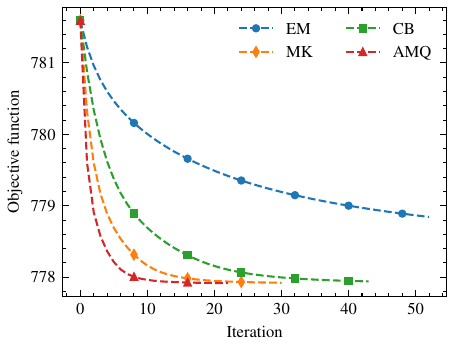}
    \end{subfigure}
    \begin{subfigure}[b]{0.32\textwidth}
        \includegraphics[height=3.5cm]{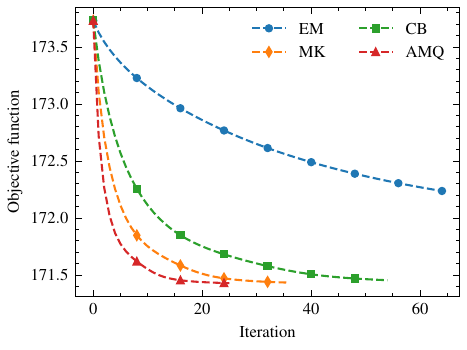}
    \end{subfigure}
    \begin{subfigure}[b]{0.32\textwidth}
        \includegraphics[height=3.5cm]{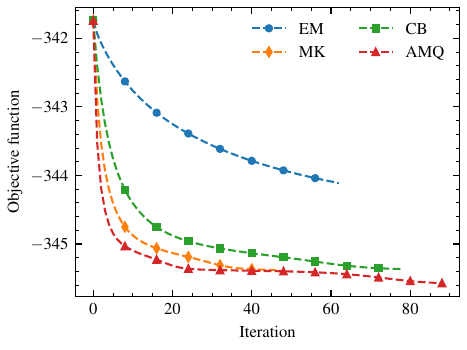}
    \end{subfigure}

    \begin{subfigure}[b]{0.32\textwidth}
        \includegraphics[height=3.5cm]{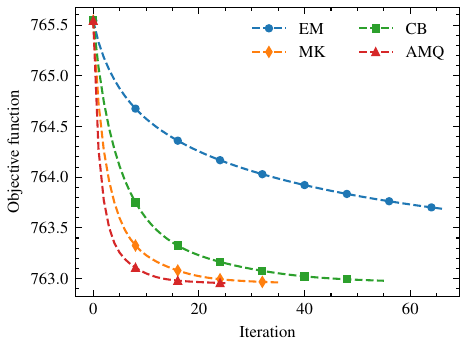}
    \end{subfigure}
    \begin{subfigure}[b]{0.32\textwidth}
        \includegraphics[height=3.5cm]{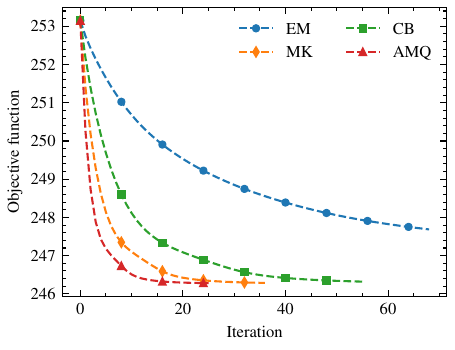}
    \end{subfigure}
    \begin{subfigure}[b]{0.32\textwidth}
        \includegraphics[height=3.5cm]{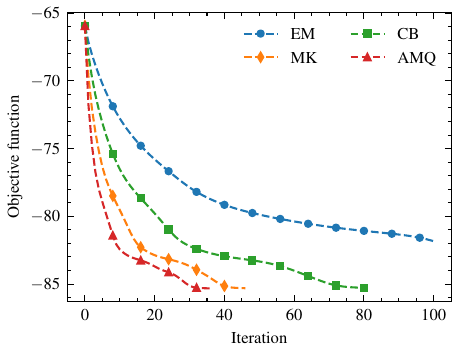}
    \end{subfigure}

    \caption{Objective function values for the Fourier reconstruction problem. Top row: \(s=10\), bottom row: \(s=80\). Left column: \(\beta=10^{-1}\), middle column: \(\beta=1\), right column: \(\beta=10\).}
    \label{fig:DCT-fxs} 
\end{figure}

For the Fourier reconstruction problem, the matrix \(\mF\) is set to be the \(256\times 512\) partial DCT matrix, where the first \(256\) rows of the \(512\times 512\) DCT matrix are selected. In this case, we do not have the true optimal solution \(\vgamma^{*}\). We will display the objective function values \(L(\vgamma^{(k)}) \) instead in \Cref{fig:DCT-fxs} for different noise levels and sparsity levels.

We observe from \Cref{fig:identity-gamma} and \Cref{fig:DCT-fxs} that the proposed algorithm converges faster than other algorithms in all the cases.

\begin{figure}[h!]
    \centering
    \begin{subfigure}[b]{0.32\textwidth}
        \includegraphics[height=3.5cm]{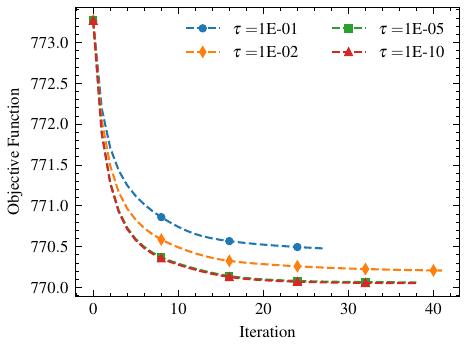}
    \end{subfigure}
    \begin{subfigure}[b]{0.32\textwidth}
        \includegraphics[height=3.5cm]{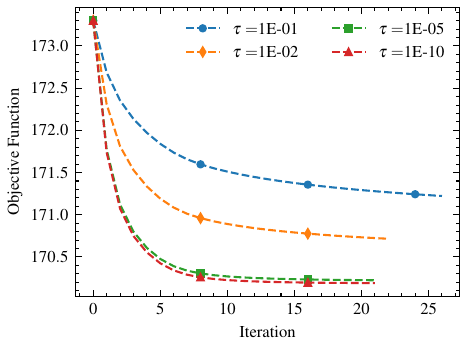}
    \end{subfigure}
    \begin{subfigure}[b]{0.32\textwidth}
        \includegraphics[height=3.5cm]{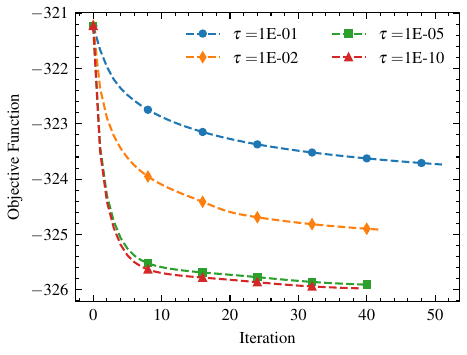}
    \end{subfigure}

    \begin{subfigure}[b]{0.32\textwidth}
        \includegraphics[height=3.5cm]{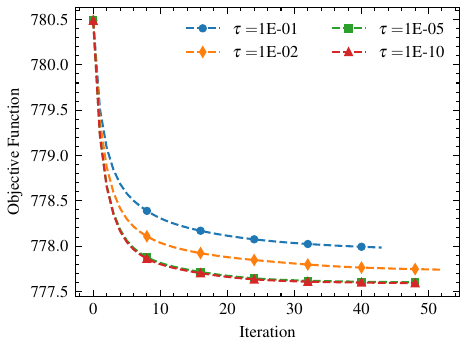}
    \end{subfigure}
    \begin{subfigure}[b]{0.32\textwidth}
        \includegraphics[height=3.5cm]{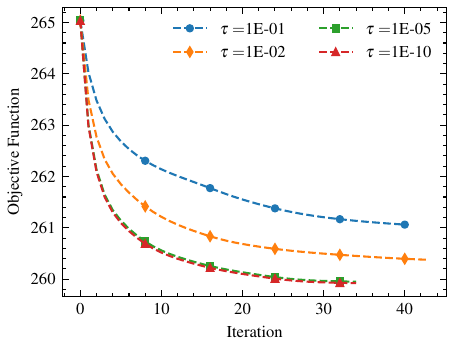}
    \end{subfigure}
    \begin{subfigure}[b]{0.32\textwidth}
        \includegraphics[height=3.5cm]{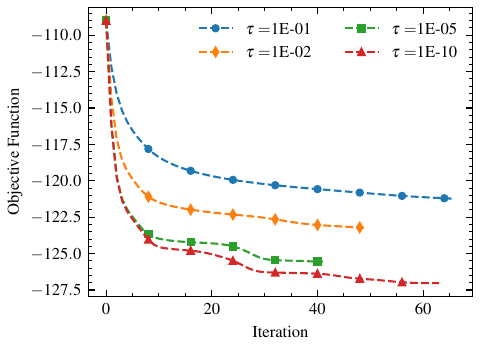}
    \end{subfigure}
    
    \caption{Objective function values for the Fourier reconstruction problem with different \(\tau\) values. Top row: \(s=10\), bottom row: \(s=80\). Left column: \(\beta=10^{-1}\), middle column: \(\beta=1\), right column: \(\beta=10\).}
    \label{fig:DCT-fxs-tau}
\end{figure}

The analysis of \(\tau\)'s influence on the convergence of the proposed algorithm presents a non-trivial challenge. Specifically, the relationship between the norm \(\left\lVert \vx^{(k+1)} - \vx^{(k)} \right\rVert\) and the parameter \(\tau\) is not immediately apparent. To shed light on this, we will conduct a series of numerical experiments. Utilizing the same partial DCT matrix and the step size choices \(\eta^{(k)}\) as previously mentioned, we examine the algorithm's performance with varying values of \(\tau\) (specifically, \(\tau=10^{-10}, 10^{-5}, 10^{-2}, 10^{-1}\)). The results, depicted in \Cref{fig:DCT-fxs-tau}, reveal a consistent trend: the convergence rate of the algorithm is accelerated with smaller values of \(\tau\). This preliminary observation lays the groundwork for a more thorough future analysis, aiming to rigorously analyze the role of \(\tau\) in the algorithm's convergence process.

\subsection{Real data: EEG and GOTCHA SAR image}
We further test the performance of the proposed algorithm on two real datasets: the EEG dataset and the GOTCHA SAR image.

Electroencephalography (EEG) is a non-invasive technique that captures brain electrical activity with high temporal resolution, serving as an indispensable tool in both basic neuroscience and clinical neurology. An EEG dataset, relevant to the study of alcoholism, is provided in 
\cite{zhang1995event}. This dataset encompasses $77$ individuals diagnosed with alcoholism and 45 control individuals (non-alcoholic). During the experiment, subjects were exposed to a stimulus, and voltage values were recorded from 64 channels of electrodes placed on their scalps. The measurements were conducted across $256$ time points and $120$ trials. By averaging the measurements over the $120$ trials, the data transforms into $122$ matrices, each sized $256 \times 64$ for every individual participant.

It is of scientific interest to investigate the relationship between alcoholism and the temporal and spatial patterns of voltage across various channels over time~\cite{li2010dimension,zhou2014regularized,huang2022robust}. A commonly employed and successful model for analyzing EEG datasets is the Generalized Linear Model of Matrix Regression. Huang et al.~\cite{huang2022robust} introduce the Robust Matrix Regression Estimator (RMRE), a novel approach that incorporates a rank constraint and $\ell_1$ regularization. This method offers valuable insights into the underlying structure of EEG datasets, revealing information that other matrix regression techniques, such asSRRE~\cite{zhou2014regularized} and LEME~\cite{hung2019low}, fail to capture. The coefficient matrix estimated by RMRE unveils the spatial-temporal dependence structure within the EEG data. The sparsity observed in the estimation highlights specific times and electrodes associated with alcoholism.

\begin{figure}
    \begin{minipage}[b]{0.49\linewidth}
        \centering
        \includegraphics[height=4cm]{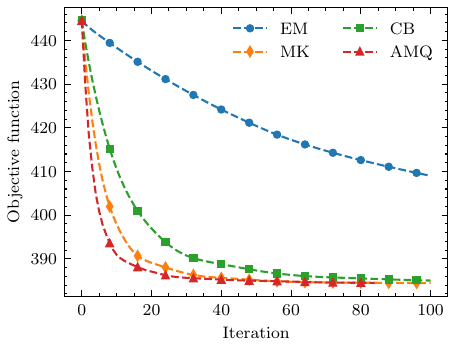}
        \caption{Convergence comparison in EEG. }
        \label{fig:converge_eeg}        
    \end{minipage}
    \begin{minipage}[b]{0.49\linewidth}
        \centering
        \includegraphics[height=4cm]{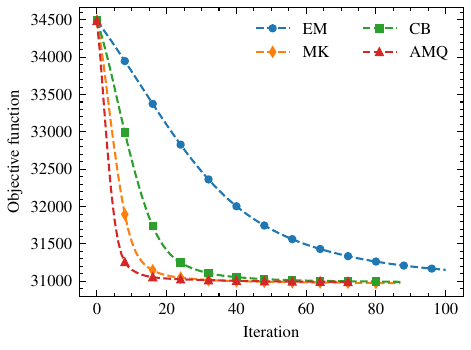}
        \caption{Convergence comparison in SAR.}
        \label{fig:sar_converg}
    \end{minipage}
\end{figure}

We set the vectorization of the matrix coefficient as the sparse signal $\xbm$ and the EEG dataset as the feature matrix $\mF$. The dimension of $\xbm$ is $64\times 256 = 16384$ and there are $590$ nonzero coefficients. The entries of $\xbm$ is scaled to $[-1,1]$. The feature matrix $\mF$ is of size $122\times 16384$. The observation $\ybm = \mF\xbm+\bm{\epsilon}$ is generated with $\SNR=20$ noises. We test the performance of the proposed AMQ algorithm with the same setting \(\tau=10^{-10}\) and \(\eta^{(k)}\) in \cref{eq:eta-k}. The comparison of the convergence of the objective function is shown in \Cref{fig:converge_eeg}. We observe the same superior performance of the proposed algorithm in the EEG dataset.

We further test the performance of the proposed AMQ algorithm \cref{eq:AM-SCA} on a synthetic aperture radar (SAR) problem. Spotlight mode airborne SAR is extensively used for surveillance and mapping purposes in remote sensing due to its ability to provide all-weather day-or-night imaging. To evaluate the performance of our algorithm on a large-scale image, we consider a GOTCHA SAR image from \cite{SAR}. We resize the image to $128\times128$, resulting in $16,384$ parameters. The overall sparsity of the image is approximately $10\%$, and we employ the partial DCT matrix of size $4,096\times 16,384$ as the dictionary matrix. We add noise with an SNR  of 20db to the observations. We test the performance of the proposed AMQ algorithm, the EM algorithm, the MK algorithm, and the CB algorithm in hyperparameters estimation. The comparison of the convergence of the objective function is shown in \Cref{fig:sar_converg}. We observe the same superior performance of the proposed algorithm in the SAR image.

\section{Conclusion}\label{sec:conclusion}
In this paper, we have developed a unified AML framework for estimating hyperparameters in SBL models. Through the unified AML paradigm, we have successfully integrated existing algorithms like the EM, MK, and CB algorithms. This integrative approach not only provides a deeper understanding of these algorithms but also aids in their comparative analysis. We show that all of these algorithms could be viewed as linearized approximations of the log determinant function through various ways of linearizations. This also motivates a new algorithm with the AML framework through a different linearization of the log determinant function. We show that the proposed AML algorithm is superior to the existing algorithms when the signal-to-noise ratio is low. 

We further propose an AMQ algorithm through adding a quadratic term to enhance the convergence of the proposed AML algorithm. We show the convergence of the proposed AMQ algorithm and demonstrate its superior performance in numerical experiments with both synthetic data and real data. In particular, we show that the proposed AMQ algorithm is more efficient than the existing algorithms across diverse settings of different noise levels and sparsity levels.

Future work could focus on more refined convergence analysis of the proposed AMQ algorithm including its convergence rate and the influence of the regularization parameter \(\tau\) on the convergence. It would also be interesting to extend the proposed AMQ algorithm to other hierarchical Bayesian models with non-Gaussian noise and/or non-Gaussian priors.

\bibliographystyle{siamplain}
\bibliography{optimizationSBL,ref-SBL}

\end{document}